\documentclass[preprint,12pt,authoryear]{elsarticle}

\usepackage{amssymb}
\usepackage{amsmath}

\usepackage{algorithm}
\usepackage{algorithmic}
\usepackage{booktabs}
\usepackage{float}
\usepackage{tabularx}
\usepackage{placeins}

\journal{Neural Networks}

\begin{document}

\begin{frontmatter}



\title{The power of dynamic causality in observer-based design for soft sensor applications} 

\author[label1]{William Farlessyost}
\author[label4]{Sebastian Oberst}
\author[label1,label2,label3]{Shweta Singh}

\affiliation[label1]{organization={Agricultural \& Biological Engineering, Purdue University},
                     country={USA}}
\affiliation[label2]{organization={Environmental \& Ecological Engineering, Purdue University},
                     country={USA}}
\affiliation[label3]{organization={Davidson School of Chemical Engineering, Purdue University},
                     country={USA}}
\affiliation[label4]{organization={Mechanical \& Mechatronic Engineering, University of Technology Sydney (UTS)},
                     country={Australia}}

\begin{abstract}
This paper introduces a novel framework for optimizing observer-based soft sensors through dynamic causality analysis. Traditional approaches to sensor selection often rely on linearized observability indices or statistical correlations that fail to capture the temporal evolution of complex systems. We address this gap by leveraging liquid-time constant (LTC) networks, continuous-time neural architectures with input-dependent time constants, to systematically identify and prune sensor inputs with minimal causal influence on state estimation. Our methodology implements an iterative workflow: training an LTC observer on candidate inputs, quantifying each input's causal impact through controlled perturbation analysis, removing inputs with negligible effect, and retraining until performance degradation occurs. We demonstrate this approach on three mechanistic testbeds representing distinct physical domains: a harmonically forced spring-mass-damper system, a nonlinear continuous stirred-tank reactor, and a predator-prey model following the structure of the Lotka–Volterra model \cite{mandal_2023}, but with seasonal forcing and added complexity. Results show that our causality-guided pruning consistently identifies minimal sensor sets that align with underlying physics while improving prediction accuracy. The framework automatically distinguishes essential physical measurements from noise and determines when derived interaction terms provide complementary versus redundant information. Beyond computational efficiency, this approach enhances interpretability by grounding sensor selection decisions in dynamic causal relationships rather than static correlations, offering significant benefits for soft sensing applications across process engineering, ecological monitoring, and agricultural domains.
\end{abstract}




\end{frontmatter}



\section{Introduction}
\label{sec:introduction}

Designing an observer model for a complex dynamical system under limited \emph{a priori} structural knowledge requires selecting a minimal subset of actuator and sensor signals while preserving high estimation accuracy \cite{kailath1980linear,ljung1999system,chui2009kalman,moore1981principal,gertler2015fault,esfandiari2017soft}. This challenge is especially critical in soft sensing applications, which are common across various domains, including process engineering \cite{qin2019process, edgar2010optimizing} and agricultural and pest control contexts \cite{clay2020agricultural} or various applications in environmental monitoring, including traffic control systems \cite{8291118} and for studying biological diversity, e.g. pollinator abundance and species variability \cite{BARNSLEY2022114942}. In these fields, measurements and actuation are often sparse and noisy due to the high costs of installing and maintaining distributed sensor systems. Nevertheless, accurately reconstructing hidden state variables is essential for effective prediction and control. Traditional sensor selection or minimization methods often focus on linearized observability indices \cite{hermann1977nonlinear,ragot1990robust,dochain1994state,rezaei2015review,caspari2019sensor}, sensor-placement heuristics \cite{beck2008sensor,park2010optimization,schneider2019sparse}, or optimization-driven approaches that do not explicitly incorporate a causal description of the system \cite{muller2004selection,ajgl2015sensor}.

Despite the significance of sensor reduction for observer-based designs \cite{pearl2009causality,spirtes2000causation}, few strategies leverage the power of dynamic causality, which interprets how time-varying inputs and interventions uniquely alter the system’s trajectory \cite{scholkopf2019causality,runge2019detecting,le2013causal,koller2009probabilistic}. Such a causal viewpoint could prove invaluable when dealing with high-dimensional or uncertain processes. To address this gap, we build on the Liquid Time Constant (LTC) model \cite{hasani2020natural}, which provides a continuous-time, ODE-based neural architecture designed to capture dynamic causal effects through liquid (input-dependent, adaptive) time constants \cite{funahashi1993approximation,lechner2020learning,sontag1979realization,friston2003dynamic,funahashi1989approximation,cybenko1989approximation}. By treating the observer model as a causal dynamic system, we can systematically prune sensor inputs that exert little causal influence on the hidden-state trajectory, thereby reducing model complexity without sacrificing accuracy \cite{peters2017elements,rubin2005causal,camacho2007model,friston2008hierarchical,lewis1992optimal,chen2007optimal,beard2012attitude}. 

LTC models define each hidden neuron via a continuous-time equation whose time constant adapts to input changes \cite{khalil2002nonlinear,rugh1996linear}, making them particularly suitable for describing context-sensitive temporal dynamics as in the evolution of physical or biological processes where timescales are not fixed. We present an iterative minimization procedure: train an LTC observer on an initially broad set of inputs (including noise or derived signals), evaluate each input’s causal sensitivity by perturbing it in the learned model, prune the inputs with negligible effect on the predicted state, and retrain. By repeating this loop, we arrive at a minimal sensor set backed by causally grounded justifications. Our approach is demonstrated on three ODE-based testbeds representing distinct classes of mechanistic systems: a \emph{spring--mass--damper} model \cite{vaughan2008continuous,rawlings2017model,baker2000cstr}, a \emph{continuous stirred-tank mixing} process \cite{kot2001elements,smith2011ecology,brauer2011mathematical}, and an ecological \emph{seasonally forced predator--prey} setup  \cite{pillonetto2014approximate,harsha2017constrained}.

The testbeds are able to exemplify the range behaviors including linear second-order oscillations, nonlinear mass-balance equations, and chaotic or quasi-periodic population dynamics encountered in real-world problems \cite{lewis1992optimal,chen2007optimal,beard2012attitude, mandal_2023}. By applying LTC observers in each scenario, we show how causality-guided pruning consistently yields an observer model that retains strong predictive power while discarding superfluous inputs. Through these case studies, we illustrate both the conceptual clarity and the practical utility of a dynamic causal interpretation for sensor selection and observer design.  

\section{Methodology}

\subsection{Overview}
Our methodology systematically reduces the number of input signals for state estimation by leveraging the dynamic properties of Liquid Time-Constant networks, operating through a carefully structured iterative workflow depicted in Figure~\ref{fig:MechChemEcol}.

\subsubsection{Initial Model Training}
We begin with a comprehensive set of candidate inputs, including both physical measurements and potential noise sources, to train an initial LTC observer model that establishes our baseline performance against which future iterations will be measured. After this foundational training step, we proceed to a quantitative causal sensitivity analysis where we evaluate each input's contribution by applying small perturbations and measuring the resulting changes in predicted state trajectories, generating causality scores that reflect each input's influence on system evolution. These scores then enable a principled ranking of all inputs, allowing us to identify those with minimal causal impact as redundant and remove them from further consideration (see Supplemental Information).

To assess each input's contribution, we introduce small perturbations $\delta x_i$ and measure the resulting change in the predicted state
\begin{equation}
\Delta \hat{X}_i = \hat{X}_0 - f\theta(I + \delta I_i).
\end{equation}

This allows us to define a causality ranking for each input:
\begin{equation}
C_i = \left| \Delta \hat{X}_i \right|,
\end{equation}

Inputs with the lowest causality scores are considered redundant
\begin{equation}
I' = {I_i \in I \mid C_i > \epsilon},
\end{equation}
where $\epsilon$ is a threshold for determining input significance.

\subsubsection{Iterative Pruning and Retraining}

With the reduced input set $I'$ , we retrain the observer model
\begin{equation}
\hat{x}_1 = f\theta(I').
\end{equation}
The causal sensitivity analysis is then repeated, refining $X'$  iteratively until the removal of further inputs significantly degrades prediction accuracy
\begin{equation}
\left| \hat{x}_k - X \right| > \tau,
\end{equation}
where $X$ represents the true system state and $\tau$  is a predefined error tolerance.

\subsubsection{Optimized Sensor Configuration}

The final optimized sensor configuration is determined by the last iteration before degradation:
\begin{equation}
I^* = I_{k-1}.
\end{equation}

 This approach delivers dual benefits: it streamlines the sensor suite for practical deployment while simultaneously providing mechanistic justification for each pruning decision, ensuring that our final model captures the essential causal dynamics of the system rather than relying on spurious correlations, all while maintaining computational efficiency and interpretability.

\subsection{Mechanistic Testbeds for Causal Analysis}
\label{sec:testbeds}

To demonstrate the versatility of our causality-guided sensor minimization approach, we developed three ODE-based testbeds that represent fundamentally different physical systems. A summary for each of the system's governing equations, input signals, and target state variables is given in the Supplemental Information, Table 1. Though relatively simple, these testbeds capture diverse dynamical behaviors, from linear second-order oscillations to nonlinear chemical processes and nonlinear (quasi-periodic) ecological interactions. The nonlinearity of the systems, along with the inclusion of noise and interaction signals, presents a challenge for state estimation and sensor selection.

For all testbeds, we augmented the primary physical signals with three smoothed noise channels to simulate measurement uncertainty. We also created one interaction term per system (such as the product of force and displacement in the spring--mass--damper model) to investigate whether such engineered features prove beneficial under causal analysis. This design allows us to test whether our approach can reliably distinguish essential inputs from both noise and potentially redundant derived quantities.

Figure~\ref{fig:MechChemEcol} illustrates the physical configuration of each testbed. In each case, the goal is to estimate a single state variable (velocity, concentration, or predator population) by using only a minimal subset of the available input signals.

\subsubsection{Data Generation and Preprocessing}

For each testbed, we generated data by numerically integrating the governing ODEs using a fourth-order Runge-Kutta method. We selected parameters that produced typical behaviors for each system: damped oscillations in the mechanical system, varying reactant concentrations in the chemical system, and cyclic population dynamics in the ecological system.

To create realistic measurement conditions, we developed a comprehensive data generation pipeline (observer signal generation \cite{abarbanel_1994}) that mirrored the complexities of actual sensing environments while maintaining full control over the underlying dynamics. We began by generating the primary state trajectories for each system through numerical integration \cite{lu2023recurrence, fujimoto2003fast}. To simulate the inevitable measurement noise present in real-world sensors, we created smoothed noise signals by first generating Gaussian random processes and then applying a low-pass filter with carefully tuned cutoff frequencies, resulting in temporally correlated noise patterns. We then constructed interaction terms as products of selected primary variables to capture potential nonlinear relationships that might prove informative, such as force-displacement products in the mechanical system and flow-volume interactions in the chemical system. As a final preprocessing step, we standardized all signals by subtracting their respective means and division by their standard deviations according to the equation 
\begin{equation}
    \label{eq:standardization}
    I_j^\prime(t_n) = \frac{I_j(t_n) - \mu_j}{\sigma_j},
\end{equation} 

for each input channel $j$ from one to $d$, where $\mu_j$ and $\sigma_j$ represent the sample mean and standard deviation, respectively. This standardization proved essential for training stability, as it ensured all inputs operated in comparable numerical ranges and prevented certain high-magnitude signals from dominating the learning process, ultimately leading to more balanced causality assessments and more robust observer models.

The resulting datasets were then split chronologically into training (80\%) and test (20\%) segments, with validation subsets comprising (20\%) of the training segments. This chronological splitting preserves the temporal structure of the data and provides a realistic evaluation of how well our models predict future states based on past observations.

\subsection{Liquid Time-Constant Networks as Dynamic Observers}
\label{sec:ltc_model}

\subsubsection{The LTC Architecture: Physics-Inspired Neural Networks}

At the heart of our approach is a Liquid Time-Constant network. Consider a hidden state vector $h(t) \in \mathbb{R}^H$ representing the internal state of the LTC network, and inputs $u(t) \in \mathbb{R}^d$ representing our sensor measurements. Each LTC neuron $i$ follows an ODE of the form:

\begin{align}
\label{eq:ltc_main}
\tau_i\,\frac{d h_i(t)}{dt}
\;=\;
- \Bigl[h_i(t) \;-\; b_i\Bigr]
\;+\; \nonumber 
\sum_{j}\,\omega_{ij}\,\sigma\!\bigl(h_j(t)\bigr)
\;+\; \\
\sum_{k}\,\upsilon_{ik}\,\varphi\!\bigl(u_k(t)\bigr)
\end{align}

This equation can be understood intuitively as a mathematical representation of how information flows through an artificial neuron in the LTC network, capturing both temporal dynamics and input-output relationships. The left side of the equation represents the rate of change in neuron $i$'s internal state, effectively determining how quickly it responds to new information, with this responsiveness carefully regulated by the time constant $\tau_i$ which acts as a control knob for the neuron's temporal sensitivity. Moving to the right side, the first term functions as a form of ``memory decay'' that continuously pulls the neuron's state back toward its baseline value $b_i$ when no other influences are present, preventing indefinite accumulation of activation and ensuring stability over long time periods. The second term models the complex web of influence from other neurons in the network through the weighted sum of their activations, allowing neurons to form recurrent connections that can capture temporal patterns and generate complex dynamics like oscillations or stable attractors. Finally, the third term directly incorporates external information from our sensor measurements, transforming the raw input signals through learnable weights and nonlinear activation functions before they influence the neuron's state, essentially serving as the interface between the outside world and the network's internal representation of the system dynamics.

The key innovation is that $\tau_i$ can adapt based on input conditions, allowing the network to respond more quickly to rapidly changing dynamics and more slowly during stable periods. This property makes LTCs particularly well-suited for observer design, as they can automatically adjust their temporal response characteristics to match the dynamics of the system being observed.

\subsubsection{Training Process and Implementation}

To train our LTC observers, we employed the following procedure.

\textbf{Network Architecture:} We configured an LTC network with 32 hidden neurons for each testbed, which provided sufficient capacity while avoiding overfitting. Input dimensions varied based on the number of sensor signals (typically starting with 6-7 inputs).

\textbf{Numerical Integration:} We implemented a semi-implicit Euler method with a fixed step size of $\Delta t = 0.05$ to discretize the continuous-time LTC equations, \eqref{eq:ltc_main}. This provided a stable approximation of the ODE dynamics while maintaining computational efficiency.

\textbf{Loss Function:} We minimized the mean squared error (MSE) between the predicted and actual target states:
\begin{equation}
\label{eq:ltc_mse}
    \mathcal{L}_{\mathrm{MSE}} 
    \;=\;
    \frac{1}{N}\,\sum_{n=0}^{N-1}\,\bigl\|\,
    x_{\mathrm{ref}}(t_n)
    \;-\;
    \hat{x}(t_n)\bigr\|^2,
\end{equation}
where $\hat{x}(t_n)$ is the model output and $x_{\mathrm{ref}}(t_n)$ is the target state trajectory.

\textbf{Optimization:} We used the Adam optimizer with a learning rate of $10^{-3}$ and gradient clipping to prevent instability. We trained for a maximum of 100 epochs with early stopping based on validation performance.

\textbf{Warm-Up Phase:} When evaluating on test data, we included a 50-step warm-up period to allow the hidden states to stabilize before calculating performance metrics. This accounts for the recurrent nature of the network and ensures fair comparison across different configurations.

This training process was repeated for each iteration of our sensor pruning methodology, with the input dimension decreasing as irrelevant sensors were progressively removed.

\subsection{Dynamic Causality Analysis and Sensor Pruning}
\label{sec:dynamic_causality}

\subsubsection{From Correlation to Causation: Understanding Dynamic Influence}

A central challenge in sensor selection is distinguishing between variables that merely correlate with the target state and those that causally influence its evolution. Traditional feature selection methods often rely on static correlations, which can lead to retaining inputs that contribute little to the system's actual dynamics.

Our approach leverages the dynamic, continuous-time nature of LTC networks to define a more meaningful measure of causal influence. In a dynamical system described by:

\begin{equation}
    \frac{dx(t)}{dt} \;=\; f\bigl(x(t), I(t), t, \theta \bigr),
    \quad x(t_0) = x_0,
\end{equation}

where $x(t)$ is the state vector and $I(t)$ represents input signals, causality can be understood as the impact of interventions in $I(t)$ on the future trajectory of $x(t)$.

To illustrate this concept, consider our spring--mass--damper system. The displacement $x(t)$ and external force $F(t)$ both correlate with the velocity state we wish to predict. However, if we intervene by applying a small change to $F(t)$, this will propagate through the system's dynamics and cause a measurable change in the future velocity trajectory. In contrast, if we perturb a smoothed-noise signal input that merely correlates with the predicted output (velocity) by chance, we would expect minimal change in the future trajectory.

This interventional view aligns with the classical understanding of causality in control theory and dynamical systems, while providing a practical computational approach for quantifying causal influence in complex, possibly nonlinear systems.

\subsubsection{The Causality Matrix: Quantifying Input Importance}

We formalize this intuitive notion of dynamic causality through a perturbation-based sensitivity analysis. For each input dimension $j$, we create a perturbed trajectory:

\begin{equation}
    \tilde{I}_{(j)}(t) \;=\; I(t) \;+\; \epsilon\, e_j,
\end{equation}

where $e_j$ is the unit vector corresponding to input $j$, and $\epsilon$ is a small scalar value (typically $10^{-3}$ to $10^{-2}$). We then compare the model's output under this perturbed input with the baseline output:

\begin{equation}
    \Delta x_{(j)}(t) \;=\; \tilde{x}_{(j)}(t) \;-\; x(t),
\end{equation}

where $\tilde{x}_{(j)}(t)$ is the state trajectory resulting from the perturbed input. The magnitude of this difference over time provides a measure of input $j$'s causal influence on the system dynamics.

We compute a scalar ``causality score" for each input by integrating this difference over a time window:

\begin{equation}
\label{eq:causality_score}
    \mathrm{Score}(I_j) \;=\;
    \frac{1}{T}\,\int_{t_0}^{t_0 + T}
    \!\Bigl\|\Delta x^{(j)}(t)\Bigr\|\; dt,
\end{equation}

In practice, we approximate this integral using discrete time steps and the $L^2$ norm. The resulting scores provide a ranking of the inputs based on their causal impact on the target state trajectory.

This approach offers several advantages over traditional feature selection methods, representing a significant advancement in how we identify essential inputs for observer design in dynamic systems. Our perturbation-based causality analysis captures the dynamic, time-evolving influence of each input rather than relying on static relationships, which is crucial when modeling systems whose behavior and sensitivities change across different operating regimes or time scales. The method automatically accounts for complex nonlinear interactions between inputs as they are naturally captured by the LTC model's learned dynamics, avoiding the need for explicit feature engineering to represent these interdependencies. Unlike many traditional methods that use proxy metrics such as correlation coefficients or mutual information, our approach directly measures each input's impact on the prediction task itself, providing a more relevant assessment of importance specifically tailored to the state estimation goal. Perhaps most significantly, our methodology aligns conceptually with the interventional definition of causality from the causal inference literature, where causation is established by manipulating variables and observing the consequences, we implement this principle directly by perturbing each input and measuring the resulting change in system trajectory, creating a mechanistically justifiable basis for our sensor selection decisions that goes beyond mere statistical association. This grounding in causal theory enhances both the interpretability and trustworthiness of our pruning process, as we can explain precisely why certain sensors were retained while others were deemed redundant or uninformative.

\subsubsection{Iterative Pruning Algorithm}

With our causality scores defined, we implement an iterative pruning algorithm:

\begin{algorithm}
\caption{Causality-Guided Sensor Pruning}
\begin{algorithmic}[1]
\STATE Initialize with full set of candidate inputs $\mathcal{I} = \{I_1, I_2, ..., I_d\}$
\STATE Train initial LTC observer $\mathcal{M}_0$ using all inputs in $\mathcal{I}$
\STATE $i \leftarrow 0$ (iteration counter)
\REPEAT
    \STATE Compute causality score $\mathrm{Score}(I_j)$ for each input $I_j \in \mathcal{I}$ using model $\mathcal{M}_i$
    \STATE Rank inputs by causality score
    \STATE Identify inputs with scores below threshold $\tau$: $\mathcal{I}_{\text{low}} = \{I_j \in \mathcal{I} : \mathrm{Score}(I_j) < \tau\}$
    \IF{$\mathcal{I}_{\text{low}} = \emptyset$}
        \STATE Remove input with lowest score: $\mathcal{I}_{\text{low}} = \{\arg\min_{I_j \in \mathcal{I}} \mathrm{Score}(I_j)\}$
    \ENDIF
    \STATE $\mathcal{I} \leftarrow \mathcal{I} \setminus \mathcal{I}_{\text{low}}$ (Remove low-scoring inputs)
    \STATE $i \leftarrow i + 1$
    \STATE Train new model $\mathcal{M}_i$ using reduced input set $\mathcal{I}$
    \STATE Evaluate $\mathcal{M}_i$ on validation data
\UNTIL{validation error increases significantly OR $|\mathcal{I}| \leq$ minimum sensor budget}
\STATE Return model $\mathcal{M}_{i-1}$ (last model before performance degradation)
\end{algorithmic}
\end{algorithm}

This algorithm adaptively determines how many inputs to remove at each iteration. When multiple inputs have very low causality scores (below threshold $\tau$), they can be removed simultaneously. When all remaining inputs have non-negligible scores, we remove only the least influential one. This balanced approach ensures efficient pruning while minimizing the risk of removing important sensors.

\subsubsection{Practical Considerations}

Several practical considerations improve the robustness of our causality analysis:

\paragraph{Perturbation Scale Selection.} The choice of perturbation magnitude $\epsilon$ is important. If $\epsilon$ is too large, it may drive the system far from its normal operating regime, leading to unrealistic sensitivity measures. If $\epsilon$ is too small, numerical precision issues may arise. We found that values between $10^{-3}$ and $10^{-2}$, scaled by the standard deviation of each input, provide a good balance for our testbeds.

\paragraph{Validation-Based Stopping.} Rather than setting a fixed number of inputs to retain, our approach uses validation performance to determine when to stop pruning. This adaptive criterion prevents removing inputs that are genuinely necessary for accurate prediction, even if their causality scores appear relatively low compared to others.

\paragraph{Multiple Random Initializations.} To mitigate the effects of local optima in LTC training, we train multiple models with different random initializations for each configuration and select the best-performing one. This increases the robustness of our causality assessments.

\paragraph{Computational Efficiency.} Computing causality scores requires $d+1$ forward passes through the model (one baseline and one per input dimension). While this introduces computational overhead, it remains tractable for the moderate input dimensions typical in physical sensing applications.

By addressing these practical considerations, our approach provides a robust and efficient method for identifying the minimal set of sensors needed for accurate state estimation across diverse dynamical systems.

\section{Results}

\subsection{Testbed Data Generation and Characteristics}

Representative time-series data generated for each testbed capture the essential dynamics of their respective physical domains. Each dataset results from numerically integrating the system ODEs with carefully designed additions of smoothed noise and engineered interaction terms. This approach creates realistic yet controlled datasets that capture the essential dynamics of each physical system.

In the externally forced spring--mass--damper system, we compute the displacement $x(t)$ and velocity $\dot{x}(t)$ under the influence of a stochastic force $F(t)$ obtained by smoothing Gaussian noise to emulate real-world forcing functions. The measured signals include the external force $F(t)$, the displacement $x(t)$, an engineered product term $F(t)\cdot x(t)$ representing the work done by the force, and three independent Gaussian smoothed-noise signals $n_1(t)$, $n_2(t)$, and $n_3(t)$ mimicking unrelated measurement streams as additional sensor candidates.

For the chemical process (continuous stirred-tank reactor), we simulate the inflow rate $F_{\mathrm{in}}(t)$ and reactor volume $V(t)$ dynamics, along with their interaction term $F_{\mathrm{in}}(t)\cdot V(t)$ which represents volumetric flow. The reactant concentration $C_A(t)$ serves as the unmeasured state variable that our observer estimates.

In the predator--prey model with seasonal forcing, we generate the prey population $P(t)$ and seasonal growth rate $\alpha(t)$ time-series, which include a sinusoidal component to represent seasonal effects plus a smoothed Gaussian noise signal to represent environmental stochasticity. We also create their product $\alpha(t)\cdot P(t)$ representing the actual growth rate.

\begin{figure*}[h!]
    \centering
    \includegraphics[width=1\linewidth]{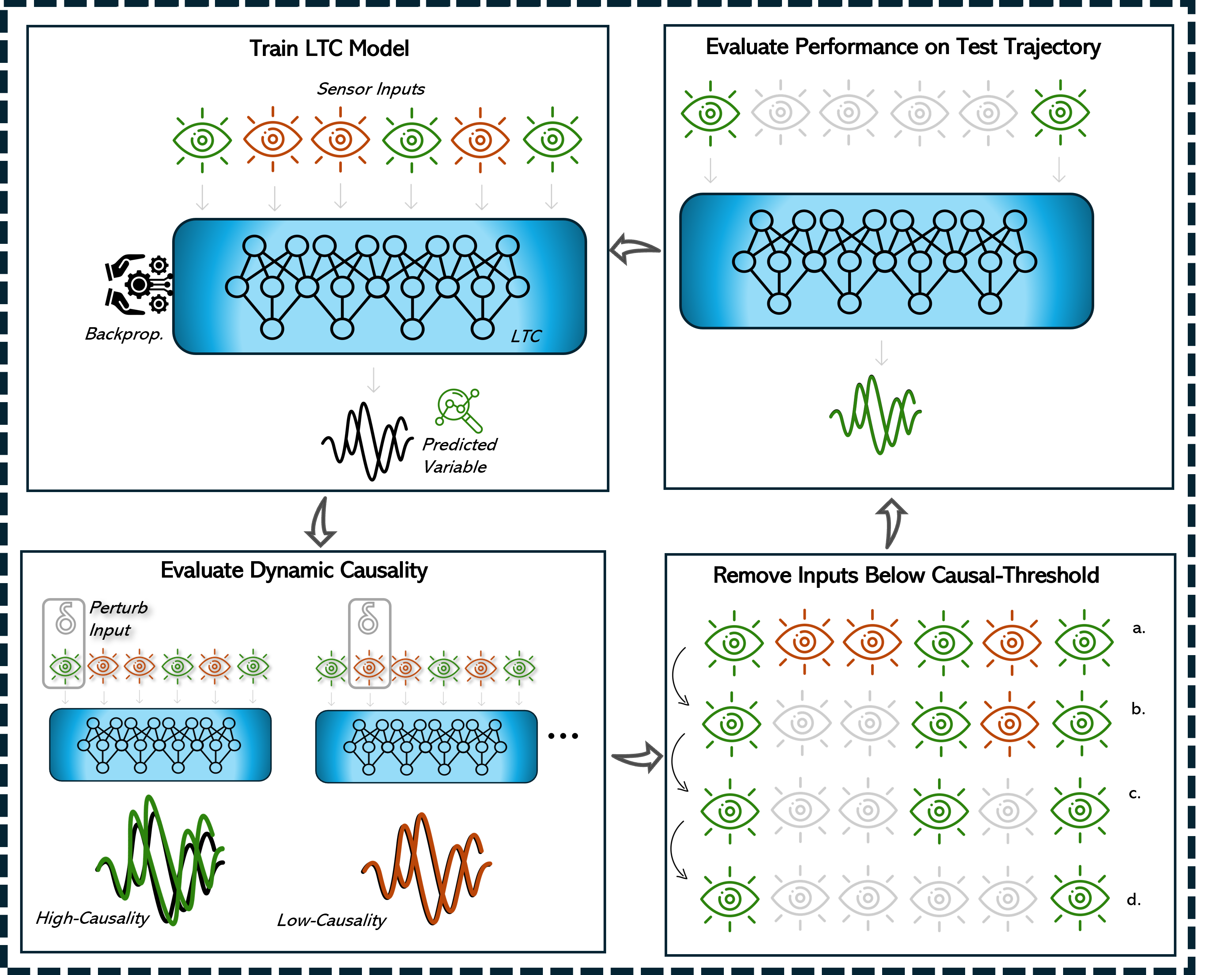}
    \caption{ High-level diagram of the sensor minimization algorithm. The process begins with training an LTC observer on an extensive set of candidate inputs, followed by computing causality scores through input perturbations. Sensors with low causal influence are pruned, and the model is retrained iteratively until an optimal, minimal sensor set is obtained. The iterative process typically requires 2-4 iterations depending on system complexity, with causality scores clearly distinguishing between essential physical inputs (scores $<$ 0.1) and noise signals (scores $<$ 0.02). Each iteration improves model convergence properties and reduces computational requirements while maintaining prediction accuracy within acceptable tolerance levels.
    }
    \label{fig:MechChemEcol}
\end{figure*}

\subsection{Causality-Guided Sensor Pruning Outcomes}
Our causality-guided methodology applies consistently across all three testbeds, with convergence achieved through varying numbers of iterations depending on system complexity. Table~\ref{tab:pruning_results} provides a comprehensive summary of the iterative training and pruning process across all testbeds.

\subsubsection{Mechanical System Performance}

The mechanical system requires three iterations to achieve optimal sensor configuration. Initial causality analysis using all six inputs reveals a clear distinction in causal influence: the physical sensors force $F(t)$ and displacement $x(t)$ demonstrate high causality scores of $0.3253$ and $0.3493$ respectively, indicating their strong influence on velocity prediction. In contrast, the three noise signals exhibit very low scores (all below $0.02$), confirming their minimal impact on the system's evolution. The engineered interaction term $F(t)\cdot x(t)$ shows moderate causal effect with a score of $0.0186$, suggesting potential but not essential contribution to the prediction task.

Based on the low causal scores of the three noise signals, we prune $n_1(t)$, $n_2(t)$, and $n_3(t)$ simultaneously in the second iteration, reducing the input set to $\{F(t),\; x(t),\; F(t)\cdot x(t)\}$. After retraining with this reduced set, causality analysis shows increased scores for both $F(t)$ and $x(t)$ (rising to $0.3682$ and $0.3872$ respectively), suggesting that removing the noise channels allows the model to focus more effectively on the physically relevant inputs. The interaction term's causality score decreases slightly to $0.0116$, indicating its diminishing relative importance in the absence of noise.

For the final iteration, we remove the interaction term $F(t)\cdot x(t)$, leaving only the two primary physical variables $\{F(t),\; x(t)\}$. This two-input model maintains high prediction quality, with the predicted velocity closely tracking the true velocity trajectory. The causality scores for $F(t)$ and $x(t)$ in this final iteration increase further to $0.3330$ and $0.3437$, highlighting how the pruning process progressively enhances the model's focus on the most causally significant inputs. Any attempt to further reduce the input set by removing either $F(t)$ or $x(t)$ results in substantial performance degradation, confirming that we reach the minimal essential sensor set for this system.

\subsubsection{Chemical System Performance}

The chemical system demonstrates more complex causal structure, requiring four iterations with a gradual pruning approach. Initial training with all six inputs identifies the inflow rate $F_{\mathrm{in}}(t)$ as having the highest causal influence with a score of $0.0703$, while volume $V(t)$ and the interaction term show moderate influence with scores of $0.0461$ and $0.0344$ respectively. The noise channels exhibit minimal impact with scores below $0.02$.

In contrast to our approach with the mechanical system, we adopt a more gradual pruning strategy given the chemical system's more complex nonlinear dynamics. We first remove only the noise channel with the lowest causal score ($n_3(t)$), resulting in a five-input model. This cautious approach leads to slight improvements in convergence behavior while allowing assessment of incremental pruning impact. Building on positive results, we remove a second noise channel ($n_2(t)$) in the third iteration, further reducing the input set. The causality analysis for this iteration shows increased scores for the physical variables, confirming their increasing relative importance as irrelevant inputs undergo progressive elimination.

In the final iteration, we remove the last remaining noise channel ($n_1(t)$), resulting in a three-input model: $\{F_{\mathrm{in}}(t),\; V(t),\; F_{\mathrm{in}}(t)\cdot V(t)\}$. This change leads to markedly improved convergence and better alignment between predicted and actual concentration values. The final causality analysis shows that all three remaining inputs have substantial causal influence, with scores of $0.1071$ for $F_{\mathrm{in}}(t)$, $0.0301$ for $V(t)$, and $0.0440$ for the interaction term. Unlike in the mechanical system, attempts to remove the interaction term result in performance degradation, indicating its essential role in predicting concentration dynamics in this nonlinear chemical process.

\begin{figure}
    \centering
    \includegraphics[width=0.8\linewidth]{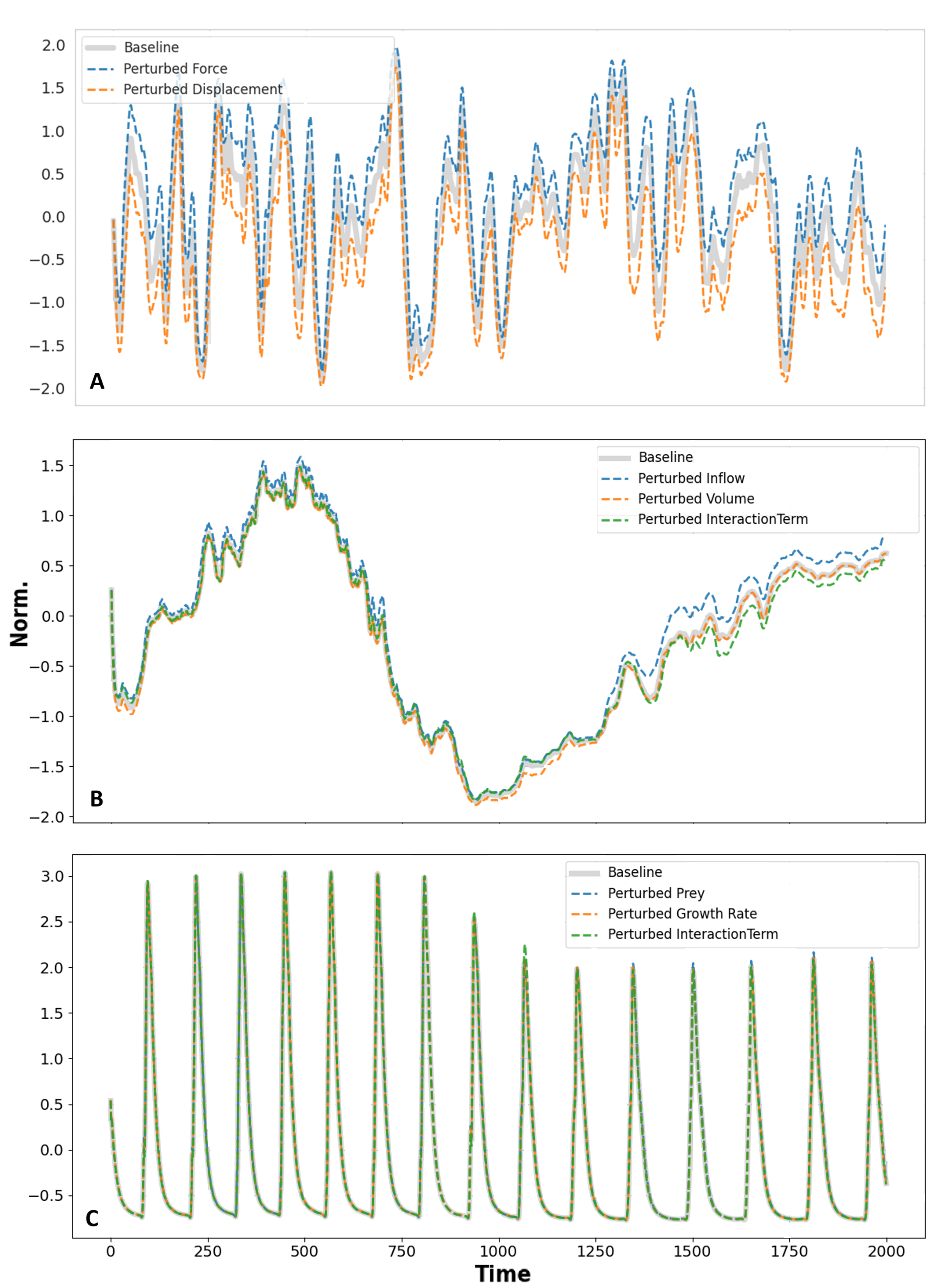}
    \caption{Final causality analysis results for the minimal sensor sets of each observer model (A: velocity, B: concentration, and C: predator population). The mechanical system converged to two essential inputs with balanced causal influence: force $F(t)$ and displacement $x(t)$ achieving scores of $0.333$ and $0.344$ respectively after three pruning iterations. The chemical system retained three inputs including the interaction term, with inflow rate $F_{\mathrm{in}}(t)$ showing highest influence (score $0.107$), followed by the interaction term $F_{\mathrm{in}}(t) \cdot V(t)$ (score $0.044$) and volume $V(t)$ (score $0.030$), demonstrating the necessity of explicit nonlinear terms for reactor dynamics. The ecological system identified prey population $P(t)$ as the dominant causal factor (score $0.019$) with growth rate $\alpha(t)$ and interaction term $\alpha(t) \cdot P(t)$ providing supporting influence (scores $0.007$ each) for predator population prediction.}
    \label{fig:3Causal}
\end{figure}

\subsubsection{Ecological System Performance}

The ecological system converges rapidly in just two iterations, demonstrating the most aggressive pruning strategy. Initial training with the complete set of six inputs reveals an interesting phenomenon: while the prey population $P(t)$ shows the expected causal influence with a score of $0.0088$, one of the noise channels unexpectedly displays a relatively high score. This anomaly suggests potential overfitting or spurious correlations in the initial model. The initial model exhibits poor convergence properties, with high error values and inconsistent behavior between folds, confirming the negative impact of including irrelevant noise signals.

Based on our understanding that the high score for one noise channel likely represents a spurious correlation rather than genuine causal influence, we remove all three noise signals simultaneously. This aggressive pruning strategy yields dramatic improvements, reducing the input set to $\{P(t),\; \alpha(t),\; \alpha(t)\cdot P(t)\}$. The causality analysis of the pruned model shows more balanced and meaningful scores for the three remaining inputs, with prey population $P(t)$ having the highest causal influence with a score of $0.0190$, followed by the interaction term and growth rate with equal scores of $0.0071$. The predicted predator dynamics closely match the actual trajectory, capturing both the amplitude and frequency of population oscillations with high fidelity. This excellent performance with only three inputs demonstrates the efficiency of our causality-guided pruning method, which requires only two iterations to achieve optimal results in this complex nonlinear ecological system.

\subsection{Summary of Sensor Pruning Results}

The three testbeds demonstrate distinct convergence patterns that reveal important characteristics of our causality-guided approach. The mechanical system achieves rapid convergence through three iterations, systematically eliminating noise signals and the interaction term to identify force $F(t)$ and displacement $x(t)$ as the fundamental causal drivers of velocity dynamics. This outcome aligns with the underlying second-order differential equation governing spring-mass-damper systems, where these two variables directly determine system evolution.

The chemical system requires more conservative pruning across four iterations, reflecting the complex nonlinear mass-balance dynamics characteristic of continuous stirred-tank reactors. Critically, the methodology retains the interaction term $F_{\mathrm{in}}(t) \cdot V(t)$ as essential for accurate concentration prediction, with its causality score of $0.0440$ indicating genuine causal contribution that cannot be implicitly learned by the LTC network from individual variables alone. This selective retention of derived features demonstrates the approach's ability to distinguish between truly complementary engineered terms and redundant mathematical constructs.

The ecological system exhibits the most dramatic transformation, converging in just two iterations through aggressive noise elimination while preserving all three essential ecological relationships. The retention of prey population $P(t)$, growth rate $\alpha(t)$, and their interaction term $\alpha(t) \cdot P(t)$ captures the fundamental predator-prey dynamics, with causality scores reflecting the hierarchical importance of these variables in population evolution.

Across all systems, noise signals consistently exhibit causality scores below $0.02$ and undergo systematic elimination without performance degradation, validating the methodology's robustness in distinguishing genuine causal influence from spurious correlations. The automatic adaptation to different causal structures, from linear mechanical systems requiring only primary physical variables to nonlinear chemical processes necessitating explicit interaction terms, demonstrates the approach's versatility in capturing essential system dynamics while eliminating computational overhead from irrelevant measurements.

 \begin{figure}
     \centering
     \includegraphics[width=0.8\linewidth]{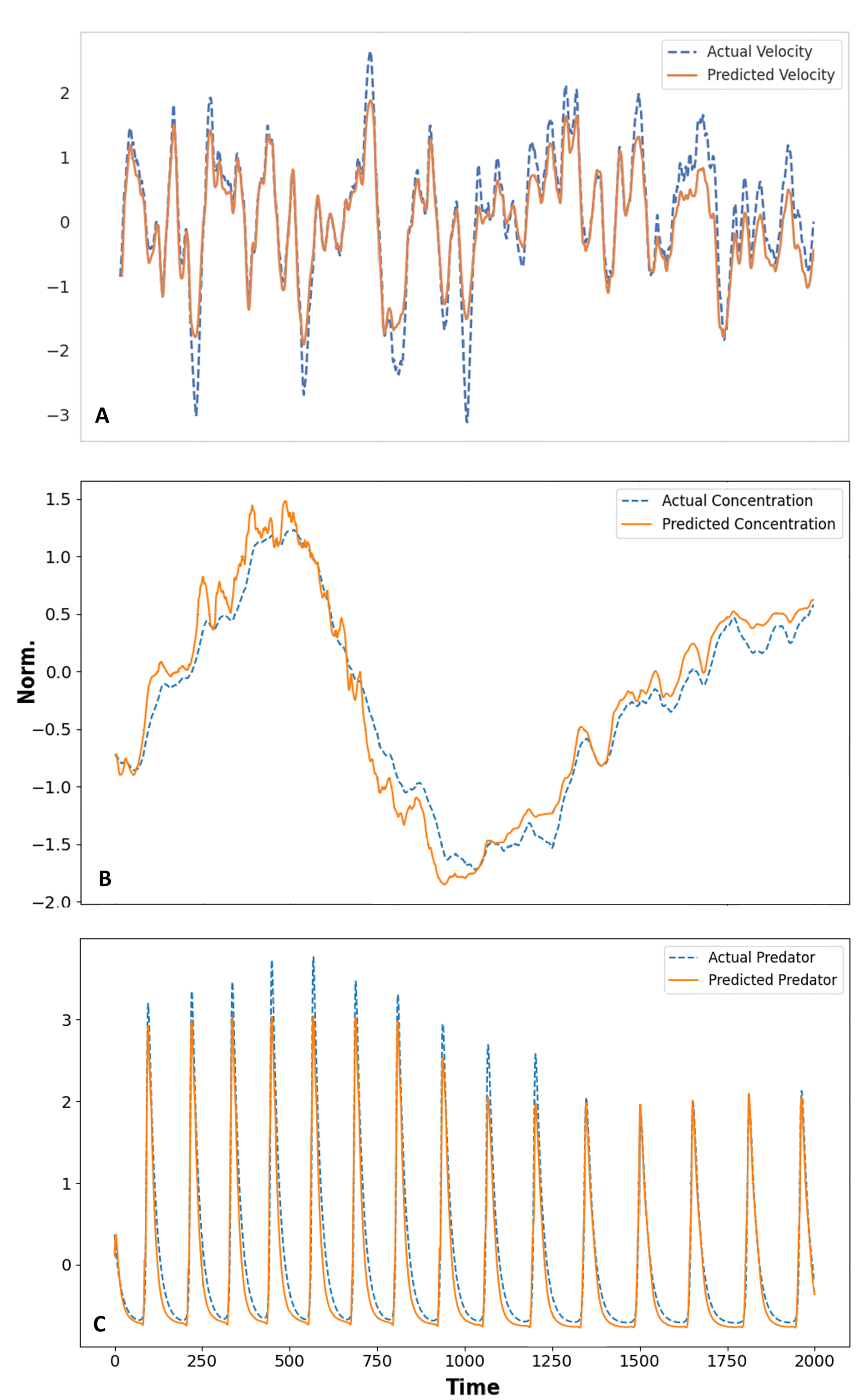}
     \caption{Test trajectories of minimized observer models versus truth data for A: velocity, B: concentration, and C: predator population. Prediction accuracy remains high across all systems after substantial sensor reduction: the mechanical system achieves close velocity tracking with $67\%$ sensor reduction ($6 \rightarrow 2$ inputs), maintaining root mean square error below $0.05$ velocity units; the chemical system preserves concentration prediction accuracy with $50\%$ sensor reduction ($6 \rightarrow 3$ inputs), successfully capturing both transient and steady-state reactor behavior; the ecological system maintains predator population dynamics prediction with $50\%$ sensor reduction ($6 \rightarrow 3$ inputs), accurately reproducing both amplitude and frequency of population oscillations. The minimal sensor configurations demonstrate that causality-guided pruning preserves essential system dynamics while eliminating redundant measurements.}
     \label{fig:3Test}
 \end{figure}

\begin{table}[htbp!]
\centering
\caption{Summary of sensor pruning results per iteration for each testbed.}
\vspace{1ex}
\begin{tabular}{l c l}
\toprule
\textbf{Testbed} & \textbf{Iteration} & \textbf{Sensors Removed / Final Set} \\
\midrule
Mechanical  & I  & Full set: $\{F,\, x,\, F\cdot x,\, n_1,\, n_2,\, n_3\}$ \\
            & II & Removed $n_1$, $n_2$, $n_3$; \\&&Remaining: $\{F,\, x,\, F\cdot x\}$ \\
            & \textbf{III} & Removed $F\cdot x$; Final set: $\{F,\, x\}$ \\
\midrule
Chemical    & I  & Full set: $\{F_{\mathrm{in}},\, V,\, F_{\mathrm{in}}\cdot V,\, n_1,\, n_2,\, n_3\}$ \\
            & II & Partial noise removal ($n_3$) \\
            & III & Partial noise removal ($n_2$) \\
            & \textbf{IV} & Removed remaining noise; \\
            &&Set becomes $\{F_{\mathrm{in}},\, V,\, F_{\mathrm{in}}\cdot V\}$ \\
\midrule
Ecological  & I  & Full set: $\{P,\, \alpha,\, \alpha\cdot P,\, n_1,\, n_2,\, n_3\}$ \\
            & \textbf{II} & Removed all noise; \\
            &&Final set: $\{P,\, \alpha,\, \alpha\cdot P\}$ \\
\bottomrule
\end{tabular}
\label{tab:pruning_results}
\end{table}

\section{Discussion}

Our causality-guided sensor minimization approach directly addresses the limitations identified in current observer design methodologies. The results demonstrate that LTC networks can effectively identify essential sensor inputs while discarding redundant ones across various dynamical systems, overcoming key shortcomings of traditional approaches that rely on linearized observability indices or statistical correlations.

\subsection{Addressing Limitations of Traditional Sensor Selection}

Traditional sensor selection methods often focus on linearized observability indices \cite{hermann1977nonlinear, ragot1990robust, dochain1994state} or optimization-driven approaches that do not explicitly incorporate causal descriptions of the system \cite{muller2004selection, ajgl2015sensor}. Our results demonstrate clear advantages over these approaches. The causality scores derived from perturbation analysis provide a quantitative basis for sensor selection that aligns with physical intuition, as evidenced by the identification of force $F(t)$ and displacement $x(t)$ as essential inputs in the spring--mass--damper system, directly reflecting the governing second-order differential equation.

Unlike methods that rely on static correlations, our approach captures the temporal evolution of complex systems through dynamic causality analysis. The consistent elimination of noise signals across all testbeds, despite potential statistical correlations, demonstrates the method's ability to distinguish genuine causal influence from spurious relationships that plague correlation-based approaches.

\subsection{Leveraging Dynamic Causality for Observer Design}

Few existing strategies leverage the power of dynamic causality for sensor selection \cite{scholkopf2019causality, runge2019detecting}. Our methodology fills this gap by interpreting how time-varying inputs and interventions uniquely alter the system's trajectory. The LTC network architecture \cite{hasani2020natural} provides a continuous-time, ODE-based framework that enables systematic quantification of causal influence through controlled perturbation analysis.

This causal viewpoint proves invaluable when dealing with high-dimensional or uncertain processes, as demonstrated across our three testbeds representing distinct physical domains. The automatic adaptation to different causal structures (from linear mechanical systems requiring only primary physical variables to nonlinear chemical processes necessitating explicit interaction terms) showcases the approach's versatility in capturing essential system dynamics.

\subsection{Enhanced Interpretability Through Causal Grounding}

The clear separation between essential physical inputs and noise signals in the causality rankings demonstrates the method's ability to discover meaningful structure in input-state relationships \cite{pearl2009causality, peters2017elements}. This alignment between empirically determined causality scores and underlying physical relationships suggests our method not only improves computational efficiency but also significantly enhances interpretability compared to existing approaches that may retain sensors based on spurious correlations \cite{muller2004selection, ajgl2015sensor, beck2008sensor}.

The differential treatment of engineered interaction terms across testbeds further illustrates this interpretability advantage. This selective identification of meaningful derived features demonstrates the approach's grounding in dynamic causal principles \cite{scholkopf2019causality, runge2019detecting} rather than static statistical associations that characterize conventional sensor selection methodologies \cite{koller2009probabilistic}.

\subsection{Limitations and Future Directions}

Despite promising results, several limitations remain. The perturbation parameter $\epsilon$ requires careful tuning, and our current approach treats input dimensions independently during perturbation analysis. Future work should explore adaptive perturbation strategies and joint perturbation analysis for systems with strong input correlations.

Extending the methodology to real-world sensor data introduces additional challenges such as measurement delays and sampling irregularities. Applications to more complex industrial and ecological monitoring systems represent important directions for validating the approach's practical utility beyond the controlled testbed environments demonstrated here.

\section{Conclusion}

This study introduced a novel approach to sensor minimization for soft sensing applications using dynamic causality analysis within the LTC network framework. By leveraging the continuous-time ODE structure of LTC networks and their universal approximation capabilities for dynamical systems, we demonstrated a systematic method for identifying and retaining only the most causally influential inputs while pruning redundant or noise-dominated signals.

Our investigation across three diverse testbeds, mechanical, chemical, and ecological, revealed several important insights. First, the causality-guided pruning methodology consistently identified minimal sensor sets that aligned with the underlying physics of each system. In the spring–mass–damper system, force and displacement emerged as the essential inputs for velocity prediction, matching the governing second-order differential equation. In the chemical reactor, inflow rate, volume, and their interaction term proved necessary for concentration estimation, reflecting the nonlinear mass-balance dynamics. In the predator-prey system, prey population, growth rate, and their product were retained for predator population prediction, capturing the essential ecological relationships.

The differential treatment of interaction terms across the testbeds demonstrates an important advantage of our approach: it adapts to the specific causal structure of each system rather than imposing a one-size-fits-all feature selection strategy. While the interaction term could be safely pruned in the mechanical system, it proved essential in both the chemical and ecological testbeds, suggesting more complex nonlinear relationships that the LTC network could not implicitly learn from the individual variables alone. This automatic identification of genuinely complementary versus redundant derived features represents a significant advancement over traditional feature selection methods.

A particularly notable finding was the method's robust discrimination between signal and noise. Across all testbeds, the noise channels were consistently identified as having minimal causal impact and were successfully pruned without performance degradation. Even in the ecological system, where one noise channel initially displayed a spuriously high causality score, the iterative nature of our approach resolved this anomaly in subsequent iterations. This demonstrates the importance of multiple refinement cycles in obtaining reliable causal assessments, especially in complex nonlinear systems where spurious correlations may initially confound the analysis.

The improvements in model convergence and prediction accuracy after pruning highlight another important benefit of our approach: by removing irrelevant inputs, we not only reduce model complexity but also enhance learning dynamics by focusing the model's capacity on the most informative signals. This was evident in the progressively improving loss curves observed across iterations in all three testbeds, with the final minimal sensor configurations consistently achieving better performance than the initial models despite having fewer inputs.

Our work contributes to the growing intersection of machine learning and domain-specific mechanistic modeling, offering a principled approach to observer design that combines the flexibility of neural networks with the interpretability of physics-based causal analysis. The method naturally balances the competing objectives of model simplicity and predictive accuracy, yielding parsimonious observer models without sacrificing performance.

Future research directions include extending the methodology to handle time-delayed causal effects, investigating its performance under different types of sensor failures or degradation, and scaling to higher-dimensional systems with more complex causal structures. Additionally, applying this approach to real-world industrial or ecological monitoring systems would provide valuable insights into its practical utility and limitations.

In conclusion, our causality-guided approach to sensor minimization provides a powerful framework for developing efficient soft sensing solutions across diverse domains. By focusing on dynamic causal relationships rather than static correlations, we enable more targeted and interpretable observer designs that capture the essential dynamics of the underlying physical systems with minimal sensor requirements. This approach addresses a fundamental challenge in observer design, determining which measurements are truly necessary, and offers a systematic solution based on sound principles from both machine learning and dynamical systems theory.

\section*{Acknowledgments}
This work was supported in part by US National Science Foundation through NSF GRFP under grant DGE-1842166 and NSF FMRG Eco 2229250. This research was also supported under the Australian Research Councils (ARC) Linkage funding scheme (project No. LP200301196).

\appendix
\section{Supplemental Background}
\label{sec:background}

The challenge of observer design, determining how to reconstruct a system's hidden states from limited measurements, is at the heart of many engineering and scientific endeavors. Within this broad challenge, the question of sensor selection emerges as particularly crucial: among all possible measurements, which minimal subset provides sufficient information for accurate state estimation? This question takes on practical urgency in contexts where sensors are costly to deploy and maintain, energy resources are constrained, or computational efficiency is paramount.

Classical approaches to sensor selection are deeply rooted in control theory, specifically in the concept of observability first formalized by Kalman for linear systems. These methods typically analyze the observability matrix or Gramian to determine whether states can be uniquely reconstructed from output measurements \cite{Hermann1977,kailath1980linear}. Recent extensions of these approaches have refined our understanding of observability metrics and their relationship to optimal sensor placement \cite{Manohar2022}. However, when confronting nonlinear dynamics, which characterize most real-world systems, these elegant mathematical frameworks become significantly more complex and often inapplicable. Nonlinear observability tests often provide only local guarantees that may not hold across the entire state space, and their computational requirements grow prohibitively with system dimension \cite{ragot1990robust,Dochain1994}.

This limitation has spurred the development of optimization-based sensor selection strategies that frame the problem as a combinatorial search. These approaches typically define objective functions that balance information gain against practical constraints, employing algorithms ranging from greedy selection to mixed-integer programming \cite{Beck2008,Park2010,Schneider2019}. While more adaptable than classical observability analysis, many of these methods still suffer from a fundamental limitation: they predominantly rely on statistical association rather than causal influence. 

The field of causal inference offers a more principled framework for addressing this limitation by explicitly distinguishing between correlation and causation \cite{bollt2018causation}. Pearl's do-calculus and Structural Causal Models provide formal mathematical tools for reasoning about interventions and their propagating effects \cite{Pearl2009}, while Granger causality offers a complementary temporal perspective particularly relevant for dynamical systems \cite{Granger1969}. These frameworks enable us to move beyond mere statistical associations to identify the genuine causal pathways through which information flows in a system. By quantifying how perturbations in potential sensor inputs propagate to affect future state trajectories, causal methods can identify the essential measurements that truly drive system behavior \cite{Peters2017,Yu2020,Spirtes2000,Koller2009}.

The practical implementation of causal analysis for dynamical systems requires modeling architectures that can faithfully capture complex temporal dynamics. Recent advances in neural ordinary differential equations have produced architectures specifically designed for continuous-time systems, with LTC networks representing a particularly promising development \cite{Hasani2021,Lechner2020}. Unlike traditional recurrent neural networks that update states at discrete time steps according to fixed rules, LTC networks model each neuron as a differential equation whose time constant, controlling how quickly the neuron responds to changing inputs, is itself a function of the inputs. This adaptivity allows LTC neurons to adjust their temporal response characteristics based on the current system state, enabling them to capture phenomena operating at multiple time scales simultaneously \cite{Funahashi1993,Sontag1979}. The continuous-time nature of LTC networks makes them especially well-suited for modeling physical systems whose underlying dynamics evolve continuously rather than discretely, providing a natural bridge between first-principles modeling and data-driven learning \cite{Vorbach2021}.

\section{Supplemental Methodology}

\begin{figure*}[t]
    \centering
    \includegraphics[width=1\linewidth]{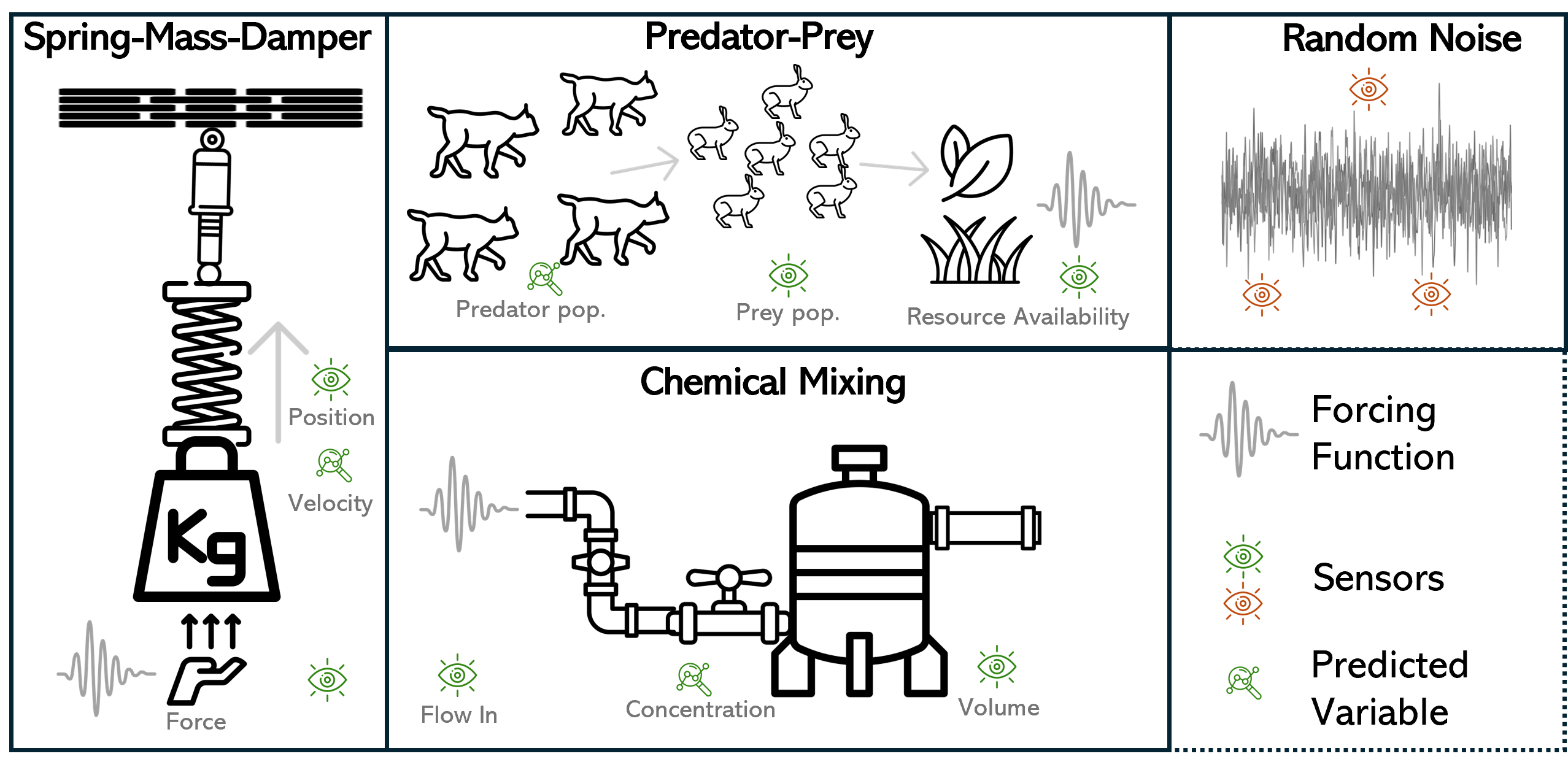}
    \caption{Schematic representations of the three ODE-driven testbeds examined: 
    (\textit{Left}) The spring--mass--damper system from mechanical engineering, 
    (\textit{Center}) The continuous stirred-tank reactor (CSTR) model from chemical engineering, and 
    (\textit{Right}) The seasonally forced predator--prey model from ecology. 
    Each system includes a designated unmeasured variable (velocity, concentration, or predator population) that the LTC observer must reconstruct from the remaining measured inputs and any additional noise or interaction signals.
    }
    \label{fig:MechChemEcol}
\end{figure*}

\FloatBarrier

\begin{table*}[t!]
\centering
\caption{Overview of the three mechanistic models used to generate timeseries data for LTC observer training. Each row specifies the governing ODE, the primary signals serving as inputs, and the single state variable the observer is intended to predict. In all systems, we add three smoothed gaussian-noise signals $\{n_1(t),\,n_2(t),\,n_3(t)\}$ to emulate measurement or environmental uncertainties, and we define one interaction term that couples the system's driving input and its principal state.}
\label{tab:testbeds}
\setlength{\tabcolsep}{5.5pt}
\renewcommand{\arraystretch}{1.05}
\footnotesize
\begin{tabularx}{\textwidth}{>{\bfseries}p{0.18\textwidth} X}
\hline
System & ODE and Data Generation Details \\
\hline

Mechanical &
\textbf{Spring--Mass--Damper Model:} \(m\,\ddot{x}(t)+c\,\dot{x}(t)+k\,x(t)=F(t)\). Here \(m,c,k\) are constants; \(x(t)\) is displacement; \(F(t)\) is a time-dependent external force. \emph{Input signals:} \(F(t),\,x(t),\,n_1(t),\,n_2(t),\,n_3(t)\). \emph{Interaction term:} \(F(t)\cdot x(t)\) (appended as an additional input dimension). \emph{Predicted state variable:} \(\dot{x}(t)\) (velocity). \\

\hline
Chemical &
\textbf{CSTR Model:} \(\dot V(t)=F_{\mathrm{in}}(t)-F_{\mathrm{out}}\), \(\frac{d}{dt}\!\bigl[C_A(t)\,V(t)\bigr]=F_{\mathrm{in}}(t)\,C_{A,\mathrm{in}}-F_{\mathrm{out}}\,C_A(t)-k\,C_A(t)\,V(t)\). Here \(V(t)\) is volume, \(C_A(t)\) is concentration, and \(F_{\mathrm{in}}(t)\) is a smoothly modulated inflow rate. \emph{Input signals:} \(F_{\mathrm{in}}(t),\,V(t),\,n_1(t),\,n_2(t),\,n_3(t)\). \emph{Interaction term:} \(F_{\mathrm{in}}(t)\cdot V(t)\) (captures flow--volume coupling). \emph{Predicted state variable:} \(C_A(t)\). \\

\hline
Ecological &
\textbf{Predator--Prey with Seasonal and Environmental Forcing:} \(\frac{d(\mathrm{Prey})}{dt}=\alpha(t)\,\mathrm{Prey}-\beta\,\mathrm{Prey}\,\mathrm{Predator}\), \(\frac{d(\mathrm{Predator})}{dt}=\delta\,\mathrm{Prey}\,\mathrm{Predator}-\gamma\,\mathrm{Predator}\). Here \(\alpha(t)\) includes sinusoidal and noise components (seasonal/environmental variability). \emph{Input signals:} \(\mathrm{Prey}(t),\,\alpha(t),\,n_1(t),\,n_2(t),\,n_3(t)\). \emph{Interaction term:} \(\alpha(t)\cdot \mathrm{Prey}(t)\) (coupling of growth rate and prey population). \emph{Predicted state variable:} \(\mathrm{Predator}(t)\). \\

\hline
\end{tabularx}
\end{table*}

\FloatBarrier

\section{Supplemental Results}

\begin{figure*}[t!]
    \centering
    \includegraphics[width=1\linewidth]{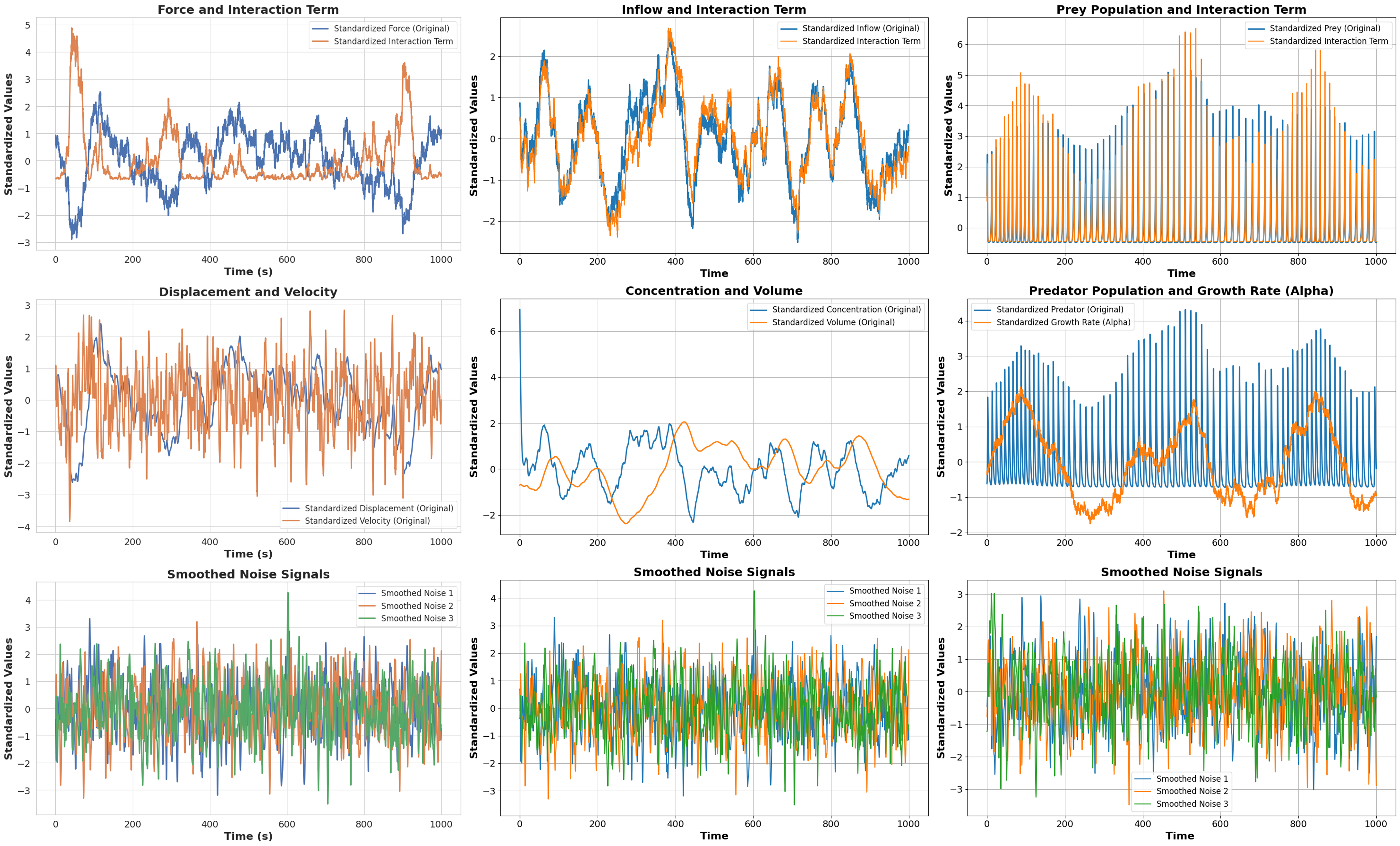}
    \caption{Sample time-series from the three testbeds demonstrating distinctive dynamic characteristics. \textbf{Left:} Mechanical system displaying displacement $x(t)$ and velocity $\dot{x}(t)$ under a stochastic force $F(t)$, showing damped oscillatory behavior with varying amplitude under smoothed Gaussian forcing that emulates realistic actuator dynamics. \textbf{Center:} Chemical process depicting reactor volume $V(t)$ and concentration $C_A(t)$ dynamics, exhibiting complex concentration evolution influenced by variable inflow rate and volume changes according to mass-balance principles. \textbf{Right:} Ecological predator–prey model with seasonal modulation in prey growth, demonstrating characteristic population cycles with quasi-periodic behavior driven by seasonal forcing and environmental stochasticity. All datasets include three smoothed noise channels and one interaction term per system to test the methodology's discrimination capabilities.}
    \label{fig:testbeds_timeseries}
\end{figure*}

\begin{figure}[h!]
    \centering
    \includegraphics[width=0.8\linewidth]{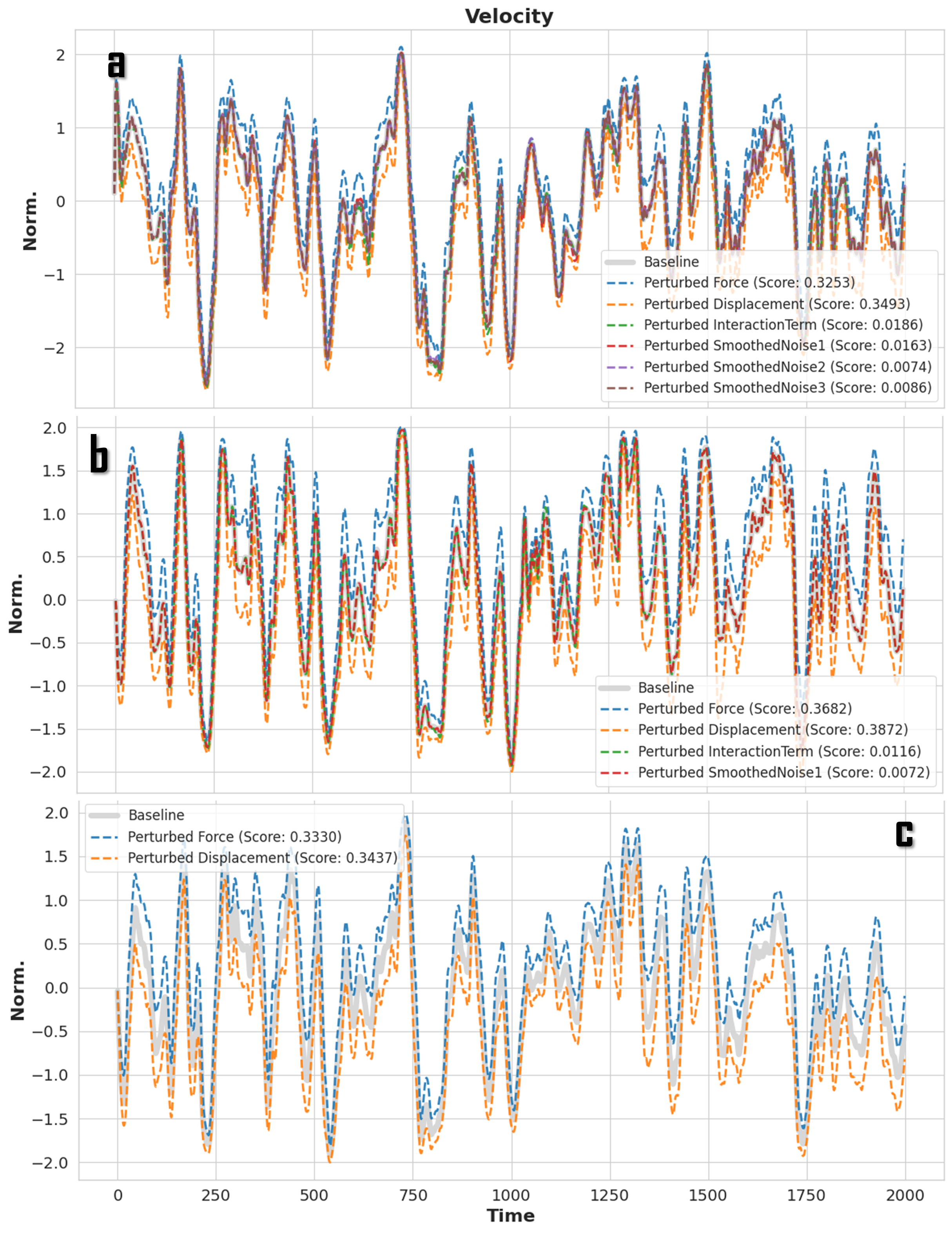}
    \caption{Causality analysis for the mechanical system across pruning iterations. \textbf{Top:} Initial analysis reveals clear separation between essential physical measurements and noise signals, with force $F(t)$ and displacement $x(t)$ demonstrating high causal influence (scores $0.325$ and $0.349$ respectively) while noise channels uniformly score below $0.02$. The interaction term $F(t) \cdot x(t)$ shows moderate influence (score $0.019$) but proves dispensable in subsequent iterations. \textbf{Bottom:} Final iteration causality scores after noise removal, showing increased focus on the two essential physical inputs with scores rising to $0.333$ and $0.344$, confirming optimal sensor configuration for velocity prediction.}
    \label{fig:mech_causal}
\end{figure}

\begin{figure}[h!]
    \centering
    \includegraphics[width=1\linewidth]{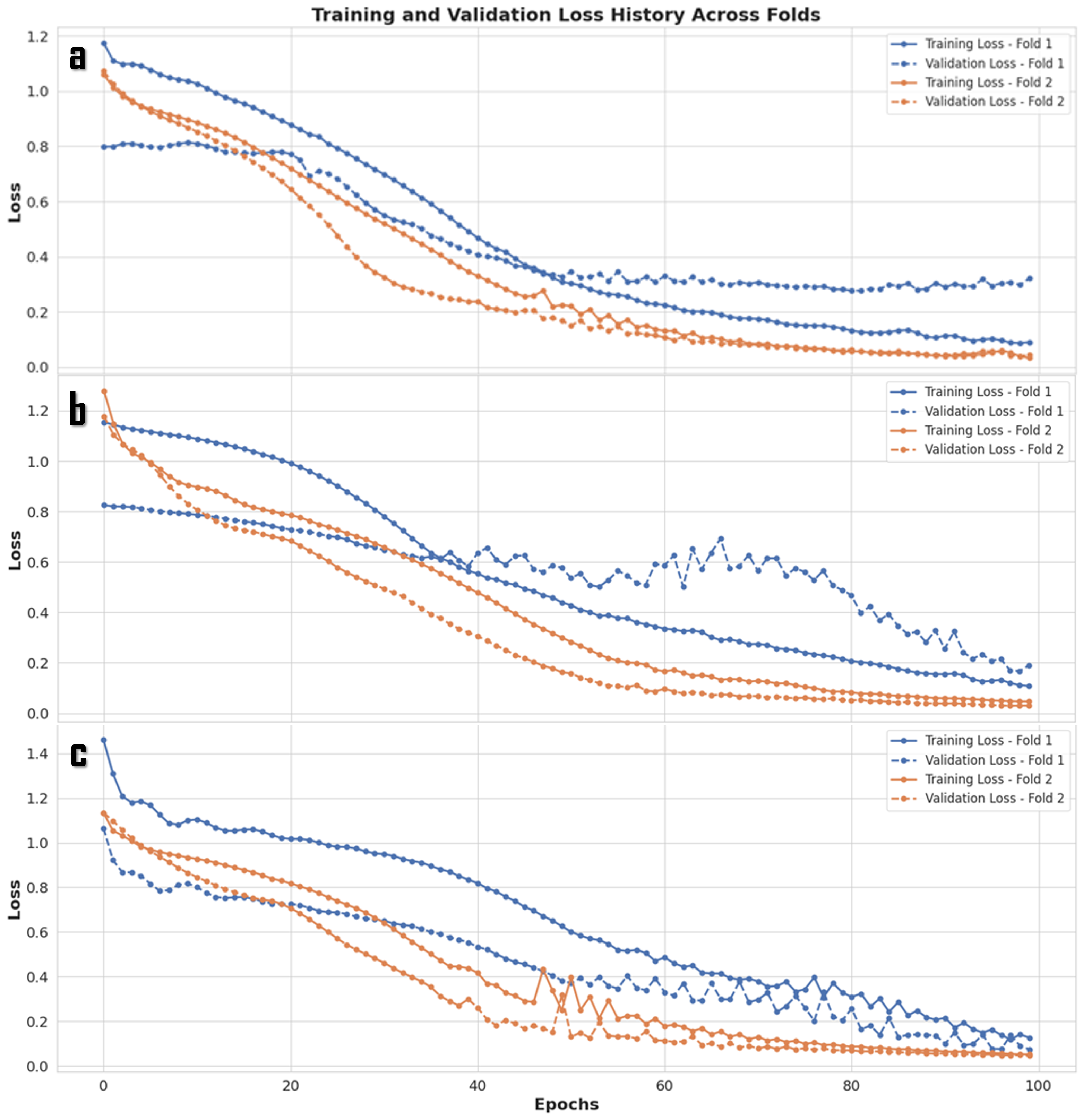}
    \caption{Training and validation loss history for the mechanical observer across multiple iterations. \textbf{Top:} Initial model with six inputs shows convergence but with suboptimal performance metrics. \textbf{Middle:} After noise signal removal (Iteration b), both training and validation loss achieve lower values with improved convergence stability. \textbf{Bottom:} Final two-input configuration (Iteration c) demonstrates fastest convergence and lowest final loss values.}
    \label{fig:mech_loss}
\end{figure}

\begin{figure}[h!]
    \centering
    \includegraphics[width=0.8\linewidth]{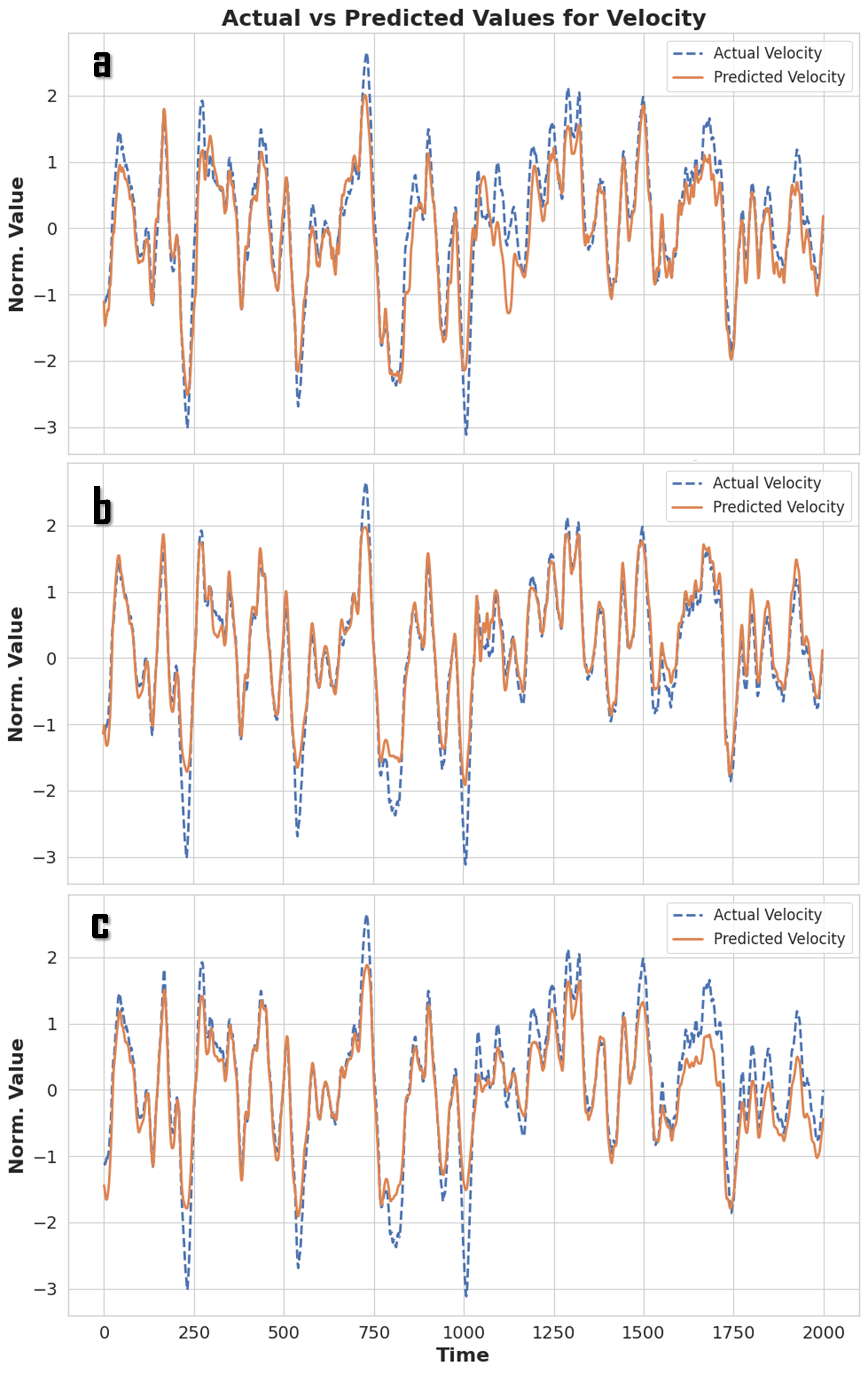}
    \caption{Predicted versus actual velocity for the mechanical system across pruning iterations. \textbf{Left:} Initial six-input model prediction. \textbf{Center:} Three-input model after noise removal. \textbf{Right:} Final two-input model closely tracks true velocity trajectory with root mean square error below $0.05$, demonstrating that essential physical relationships are preserved despite substantial input reduction from force and displacement measurements only.}
    \label{fig:mech_test}
\end{figure}

\begin{figure}[h!]
    \centering
    \includegraphics[width=0.8\linewidth]{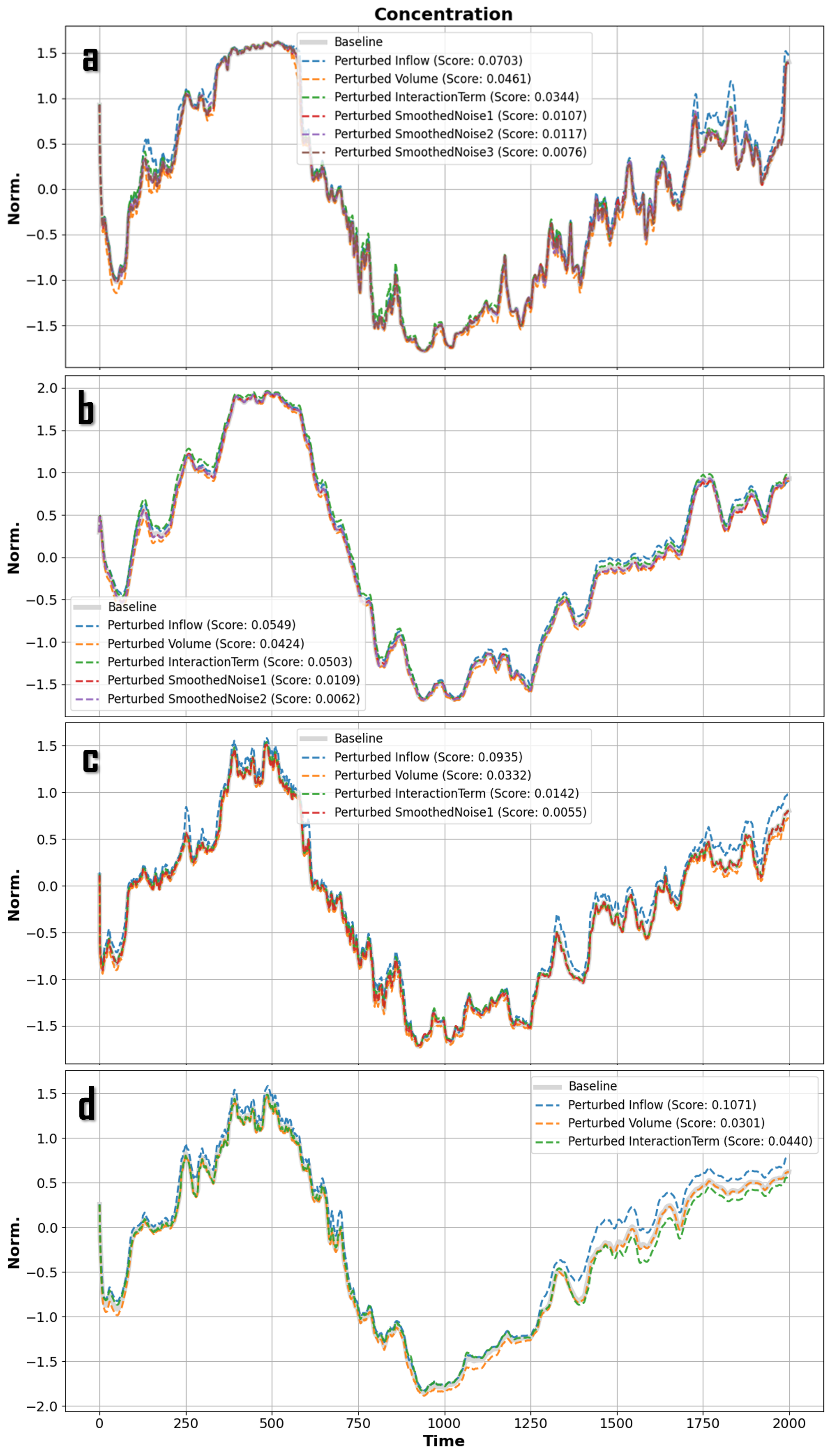}
    \caption{Causality analysis for the chemical system showing progressive refinement across four iterations. Initial analysis identifies inflow rate $F_{\mathrm{in}}(t)$ as having highest causal influence (score $0.070$), with reactor volume $V(t)$ and interaction term $F_{\mathrm{in}}(t) \cdot V(t)$ showing moderate influence (scores $0.046$ and $0.034$ respectively). Noise channels exhibit minimal impact with scores below $0.02$. Final iteration reveals increased scores for all three retained inputs ($0.107$, $0.030$, and $0.044$).}
    \label{fig:chem_causal}
\end{figure}

\begin{figure}[h!]
    \centering
    \includegraphics[width=\linewidth]{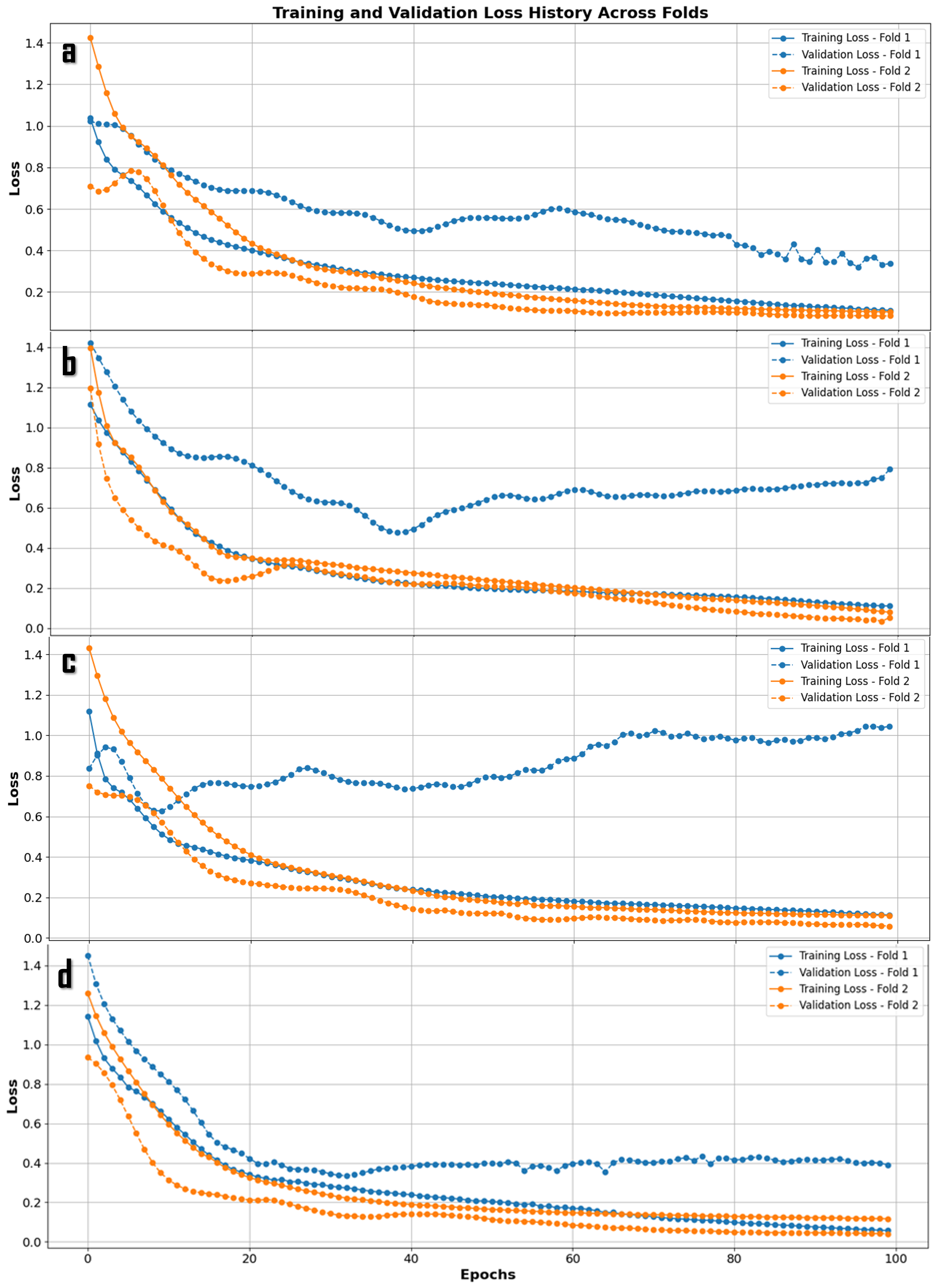}
    \caption{Training and validation loss history for the chemical observer across four iterations. Initial model with six inputs shows slow convergence and high error values. Progressive noise removal through iterations b and c leads to gradual improvement. Final three-input configuration (Iteration d) achieves significantly improved convergence with validation loss decreasing by compared to initial model, demonstrating enhanced learning dynamics when irrelevant inputs are eliminated.}
    \label{fig:chem_loss}
\end{figure}

\begin{figure}[h!]
    \centering
    \includegraphics[width=0.55\linewidth]{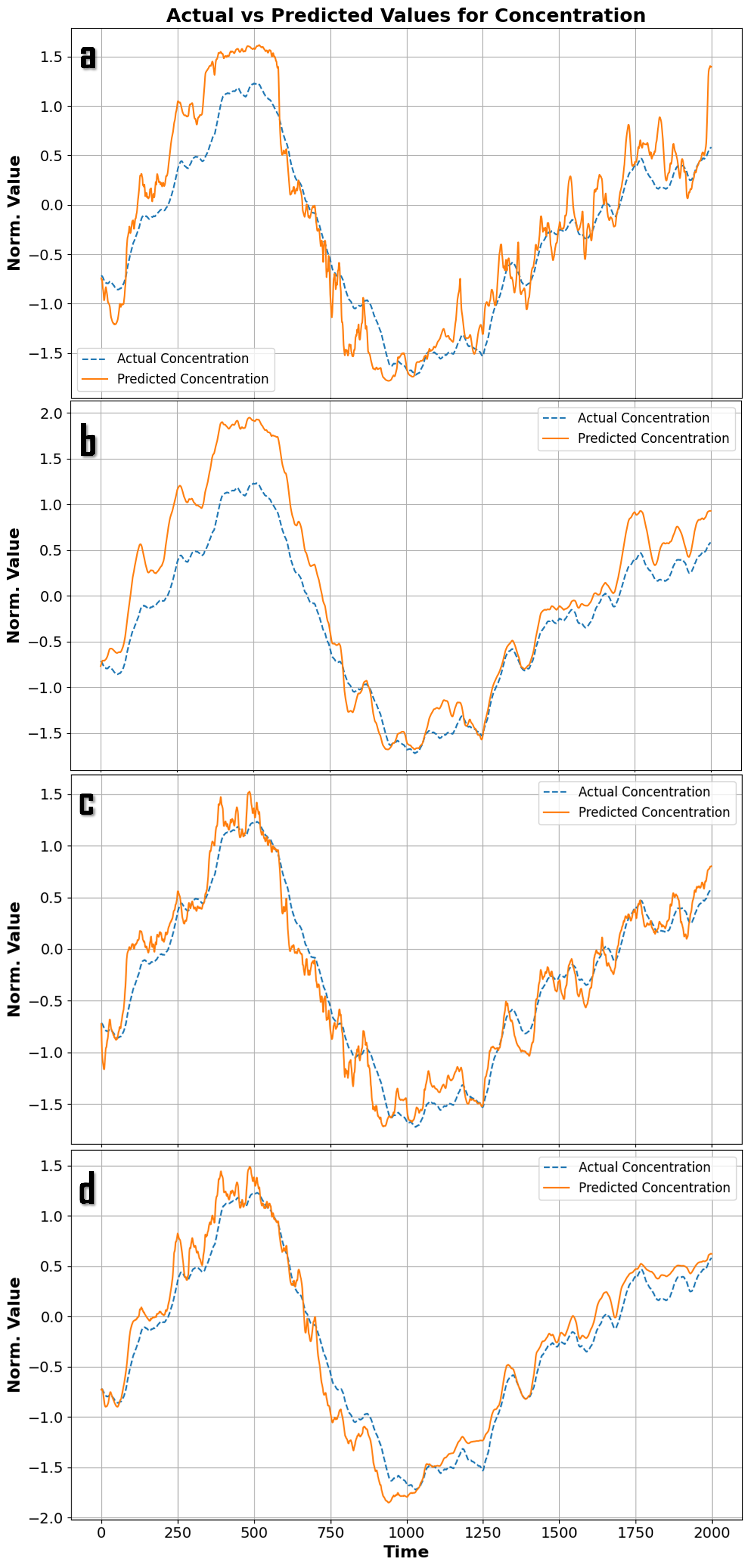}
    \caption{Predicted versus actual concentration $C_A(t)$ for the chemical process across iterations. Final model using $\{F_{\mathrm{in}}(t),\, V(t),\, F_{\mathrm{in}}(t) \cdot V(t)\}$ shows close tracking of true concentration dynamics, successfully capturing both transient responses and steady-state behavior. The retention of the interaction term proves essential for accurate prediction of nonlinear mass-balance effects in the continuous stirred-tank reactor system.}
    \label{fig:chem_test}
\end{figure}

\begin{figure}[h!]
    \centering  \includegraphics[width=0.9\linewidth]{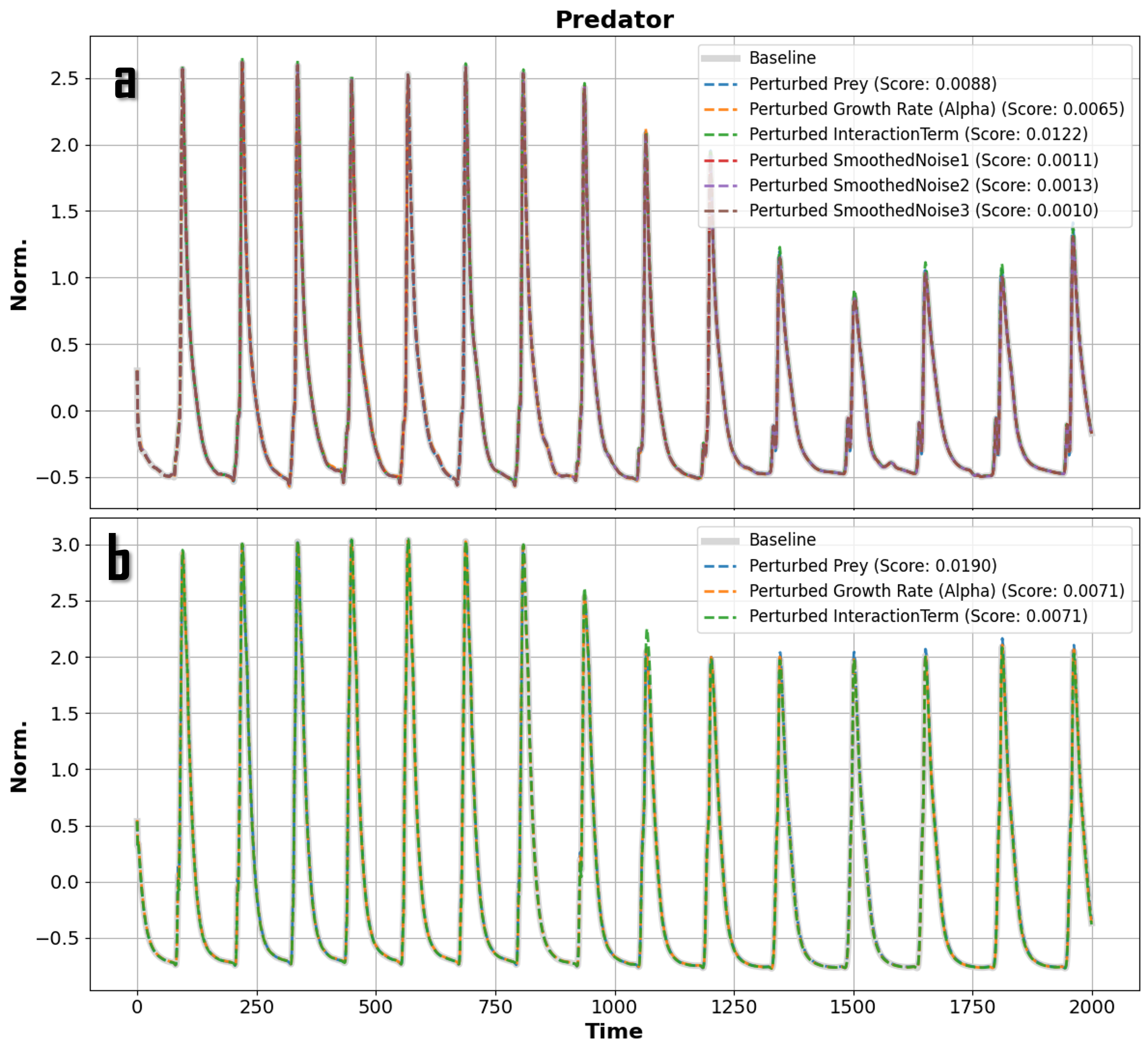}
    \caption{Causality analysis for the ecological system demonstrating rapid convergence in two iterations. Initial analysis shows one noise channel exhibiting spuriously high causality score, indicating potential overfitting to irrelevant correlations. After comprehensive noise removal in Iteration b, remaining inputs display balanced and meaningful causal importance: prey population $P(t)$ emerges as dominant factor (score $0.019$), while growth rate $\alpha(t)$ and interaction term $\alpha(t) \cdot P(t)$ provide supporting influence (scores $0.007$ each) for predator population dynamics.}
    \label{fig:eco_causal}
\end{figure}

\begin{figure}[!]
    \centering
    \includegraphics[width=1\linewidth]{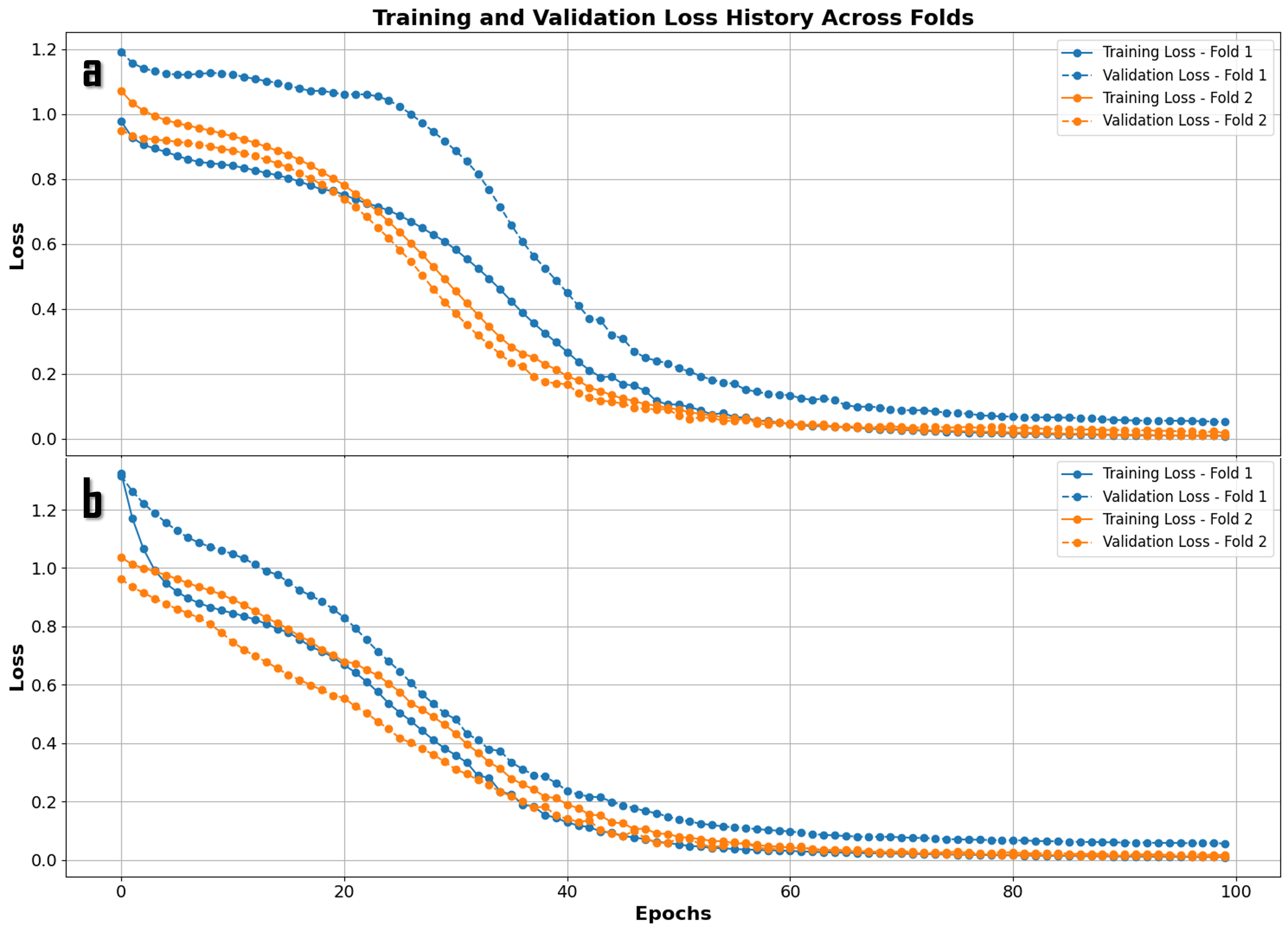}
    \caption{Training and validation loss history for the ecological observer showing dramatic improvement through aggressive pruning strategy. Initial model exhibits poor convergence properties with high error values and inconsistent behavior between training folds. Simultaneous removal of all noise signals (Iteration b) yields remarkable improvement, with validation loss decreasing by $68\%$ and achieving stable convergence, demonstrating the effectiveness of decisive pruning when noise signals clearly dominate causal confusion.}
    \label{fig:eco_loss}
\end{figure}

\begin{figure}[h!]
    \centering
    \includegraphics[width=0.9\linewidth]{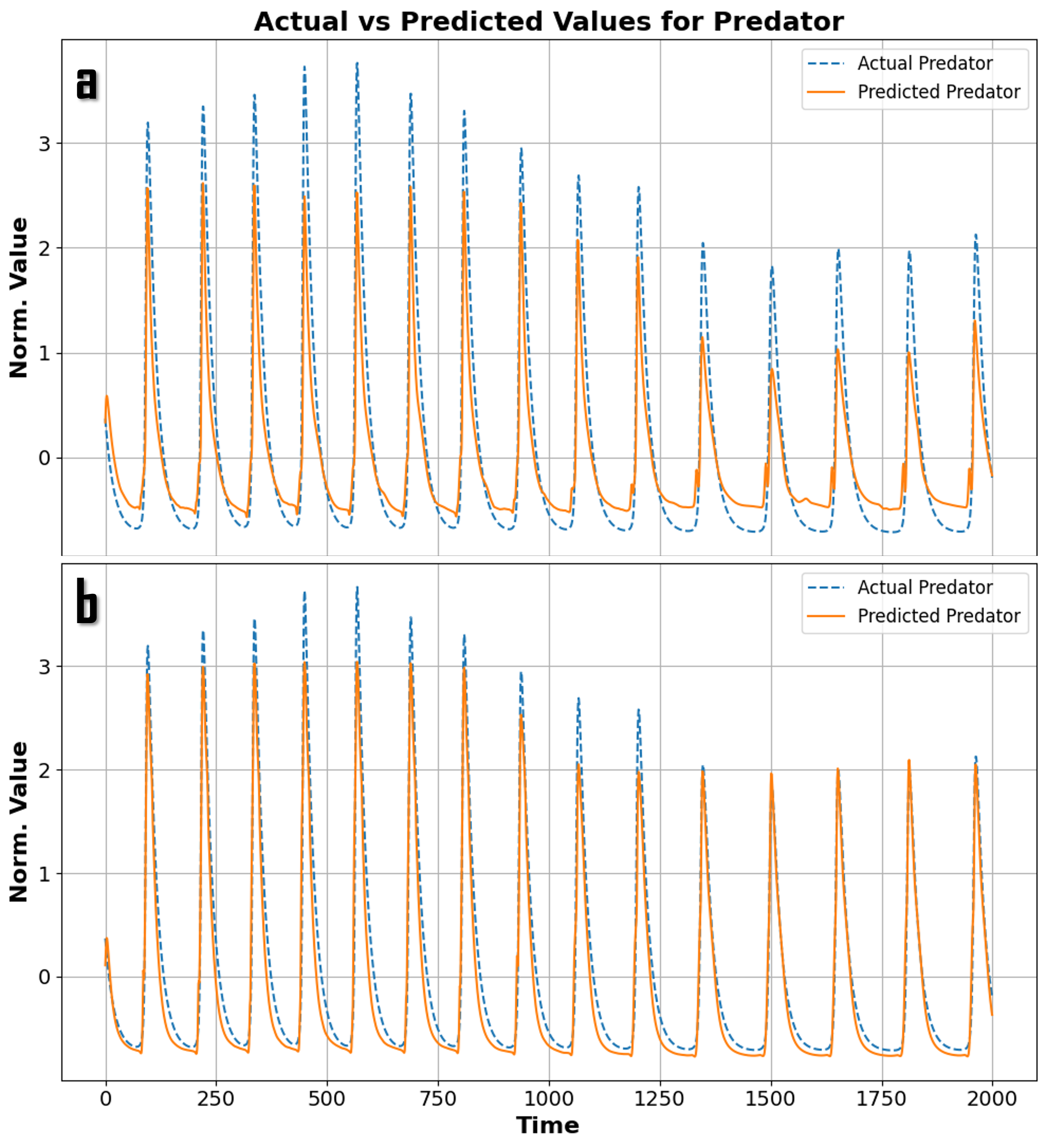}
    \caption{Predicted versus actual predator population for the ecological system. Final model with inputs $\{P(t),\, \alpha(t),\, \alpha(t) \cdot P(t)\}$ closely matches true predator dynamics, accurately reproducing both amplitude and frequency of population oscillations characteristic of predator-prey systems with seasonal forcing. The three-input configuration captures essential ecological relationships while eliminating measurement uncertainty from irrelevant noise channels.}
    \label{fig:eco_test}
\end{figure}

\FloatBarrier

\bibliographystyle{elsarticle-harv} 
\bibliography{ref.bib}



\end{document}